\documentclass[letterpaper]{article} 
\usepackage{aaai25}  
\usepackage{times}  
\usepackage{helvet}  
\usepackage{courier}  
\usepackage[hyphens]{url}  
\usepackage{graphicx} 
\usepackage{natbib}  
\usepackage{caption} 
\title{CohEx: A Generalized Framework for Cohort Explanation}
\author {
    Fanyu Meng\textsuperscript{\rm 1},
    Xin Liu\textsuperscript{\rm 1},
    Zhaodan Kong\textsuperscript{\rm 1},
    Xin Chen\textsuperscript{\rm 2}
}
\affiliations {
    \textsuperscript{\rm 1}University of California, Davis\\
    \textsuperscript{\rm 2}Georgia Institute of Technology\\
    fymeng@ucdavis.edu, xinliu@ucdavis.edu, zdkong@ucdavis.edu, xinchen@gatech.edu
}

\usepackage{subcaption}
\usepackage{xcolor}
\usepackage{booktabs}
\usepackage{enumitem}
\usepackage{amsmath}
\usepackage{amssymb}
\usepackage[ruled,linesnumbered]{algorithm2e}
\usepackage{makecell}
\usepackage{xcolor}

\begin{document}

\maketitle

\begin{abstract}
eXplainable Artificial Intelligence (XAI) has garnered significant attention for enhancing transparency and trust in machine learning models. However, the scopes of most existing explanation techniques focus either on offering a holistic view of the target model (global explanation) or on individual instances (local explanation), while the middle ground, i.e., cohort-based explanation, is less explored. Cohort explanations offer insights into the target model's behavior on a specific group or cohort of instances, enabling a deeper understanding of model decisions within a defined context. In this paper, we discuss the unique challenges and opportunities associated with measuring cohort explanations, define their desired properties, and create a generalized framework for generating cohort explanations based on supervised clustering.
\end{abstract}

%
\begin{links}
  \link{Code}{https://github.com/fy-meng/cohex}
\end{links}

\section{Introduction \& Motivation}
\label{sec:intro}

eXplainable Artificial Intelligence aims to improve the existing machine learning system by providing transparent explanations for model predictions and decisions \citep{survey2020, survey2018}. The \textit{scope} of explanation refers to the degree to which the explanation provided is generalized. The scope, as defined by \cite{fact-sheet}, is usually (1) \textit{global explanation}, which aims to explain the target model comprehensively; or (2) \textit{local explanation}, which focuses on explaining the behavior in one instance of model prediction. However, there is a less explored middle ground, namely (3) \textit{cohort explanation}, or sometimes also called \textit{regional explanation}, which generalizes the explanation on a subset of a dataset, a subspace in the model's input space or decision space. Cohort explanation can be seen as a spectrum that generalizes both local and global explanation methods, where global and local explanations are at the two extremes of conciseness and faithfulness. This generalization allows users to uncover patterns, biases, and context-specific insights that may not be apparent when examining individual instances in isolation or the model as a whole.

\begin{figure*}[t]
  \centering
  \begin{subfigure}[b]{0.3\textwidth}
    \centering
    \includegraphics[width=\textwidth]{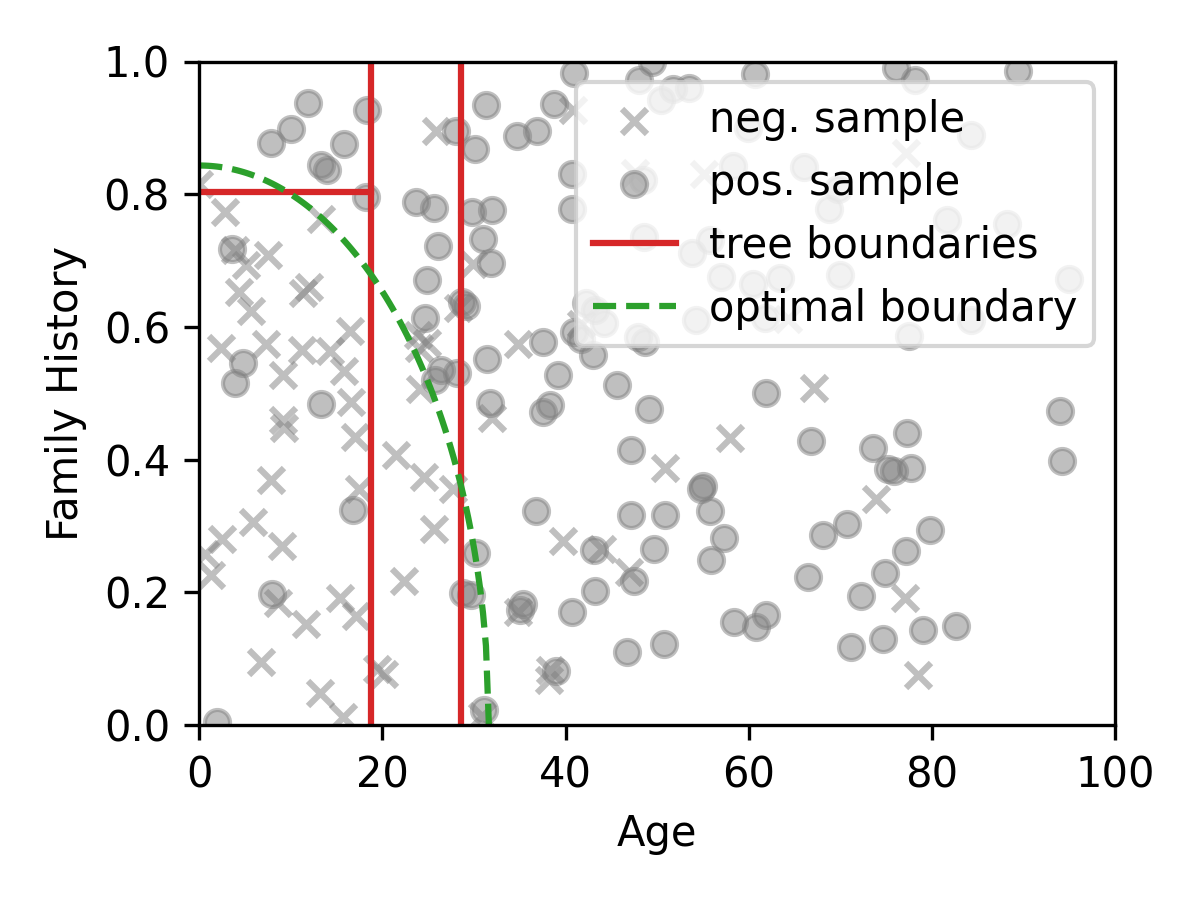}
    \caption{Data and the target decision tree.}
    \label{fig:motivation-data}
  \end{subfigure}\quad
  \begin{subfigure}[b]{0.3\textwidth}
    \centering
    \includegraphics[width=\textwidth]{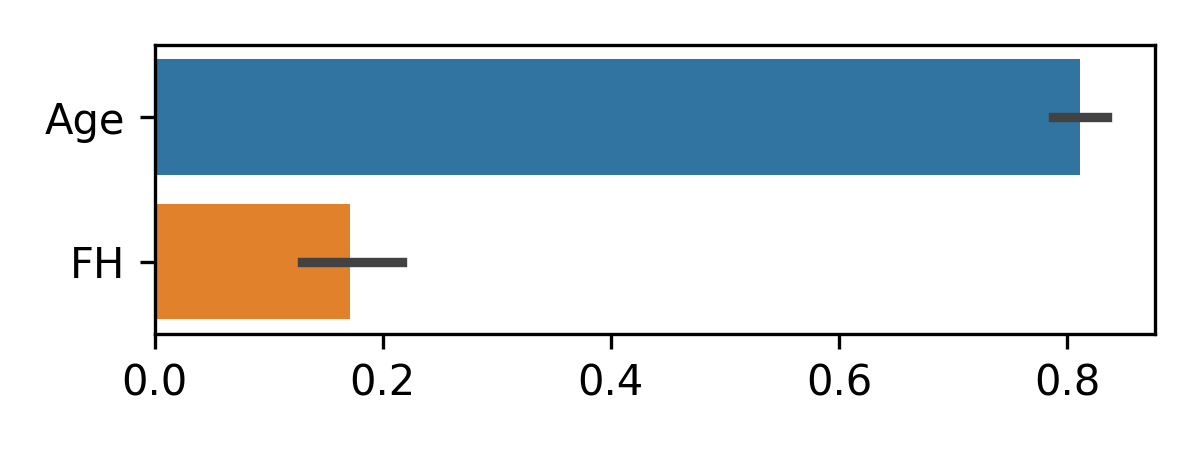}
    \includegraphics[width=\textwidth]{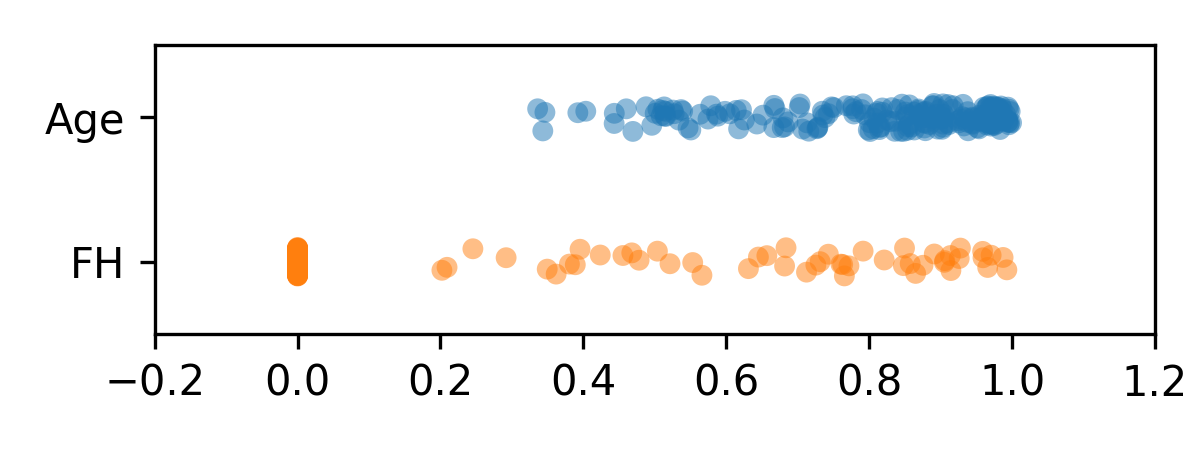}
    \caption{Global and local importance.}
    \label{fig:motivation-exp}
  \end{subfigure}\quad
  \begin{subfigure}[b]{0.3\textwidth}
    \centering
    \includegraphics[width=\textwidth]{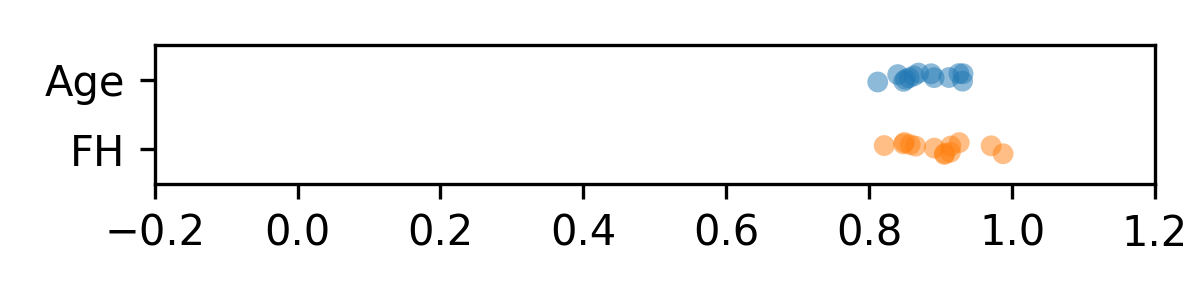}
    \includegraphics[width=\textwidth]{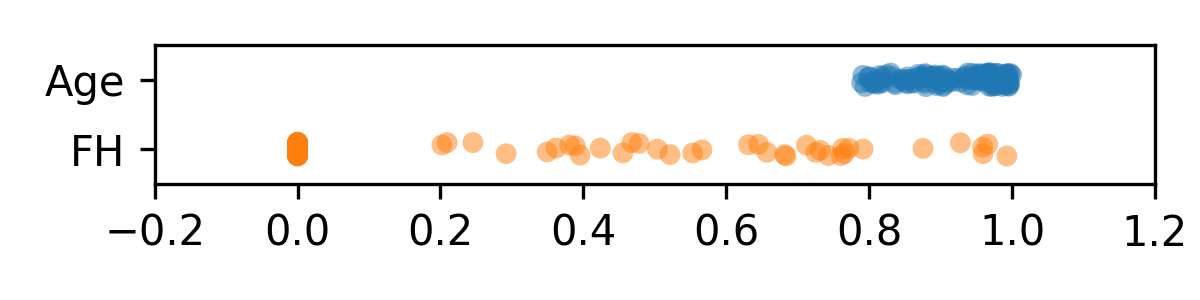}
    \includegraphics[width=\textwidth]{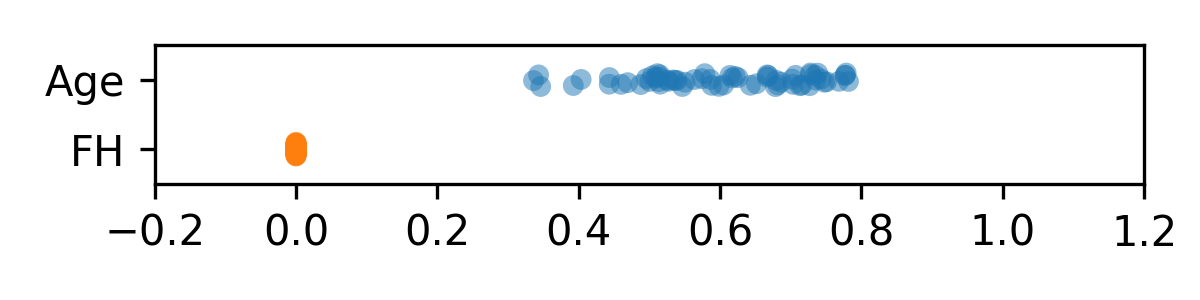}
    \caption{Cohort explanation with three cohorts.}
    \label{fig:motivation-coh}
  \end{subfigure}
  \caption{A motivating example: a synthesized patient binary classification problem.}
  \label{fig:motivation}
\end{figure*}

\paragraph{Lack of Robustness of Feature Attribution Methods } Many existing explanation methods in XAI, notably feature attribution methods, face limitations regarding robustness. The output can be sensitive to small perturbations in input or model parameters, leading to unstable or unreliable insights \cite{robustness, fooling}. Some explainer would be produce biased output under simple transformations, failing to only capture relevant information for decision-making \cite{unreliability-saliency}. A common practice to improve robustness in machine learning is via bootstrapping - averaging the output of multiple weak learners to reduce overfitting \cite{bagging, random-forests, ensemble}. A similar approach could also be beneficial for XAI. By averaging the explanations on a group of similar samples, we could reduce the noises in explanations, providing more stable, interpretable insights that account for patterns across relevant subsets of data.

\paragraph{Faithfulness vs. Conciseness }
Global XAI methods provide one general explanation, but may not generalize well on subgroups; local methods assign a specific explanation for each data sample, but it is not concise and may not help to understand the entire model. Cohort explanation methods strive for a balance between the two, with the aim of creating meaningful splittings of the feature space and assigning explanations to each cohort that generalize well within each of the groups. Thus, the cohort explanation generates ``\textit{contextful}" information: the explanation of each cohort recognizes that it can be applied to a limited region. This helps users better understand the conditions that must be maintained for the output explanation to be valid and to offer a more concise and transparent explanation \cite{fact-sheet}. 

To illustrate this motivation, consider a simplified medical patient classification task in Fig.~\ref{fig:motivation-data}. The two features are age and the fraction of family members with this disease. The likelihood of the condition occurring increases with respect to both features. The target model is a decision tree with a depth of two. Note that we treat the decision tree as a black-box, i.e. the explainer does not know that the model is a tree. 

To generate cohort explanation to the tree, we can first apply a simplified version of the counterfactual-based local explanation \cite{counterfactual}. We consider the importance of each feature as the minimum perturbation on that feature to change the prediction, and a small perturbation signals a high importance. Fig.~\ref{fig:motivation-exp} shows the global and distribution local importance on this model. These two types of explanations illustrates that in general, age has a higher importance, but they cannot used to explain whether they are patterns within the explanations. However, such pattern exists: if we separate the samples into three groups: (1) young patients with high family history, (2) middle-aged or young patients with low family history, and (3) older patients, the distribution of feature importance are distinct between the groups. Fig.~\ref{fig:motivation-coh} shows this cohort explanation, and we can provide a deeper analysis that the features are (1) both are significant; (2) only age is significant; (3) neither is significant respectively on the three groups. Naturally, this analysis correspond to the slope of the optimal decision boundary in Fig.~\ref{fig:motivation-data}. Cohort explanation simplifies the mechanisms of the model into a few regions in which different features are the main factor behind decision making. Additionally, it provides insights that cannot be provided by global or local explanations.

Our contributions in this paper are as follows: 
\begin{enumerate}[noitemsep,topsep=0pt,parsep=0pt]
  \item We propose desired properties of cohort explanation and devise their evaluation metrics;
  \item We develop a generalized framework to convert existing data-driven local feature importance methods to cohort explanations;
  \item We conduct experiments on several examples and demonstrate the superior performance of our algorithm over existing benchmarks based on the proposed metrics.
\end{enumerate}

\section{Related Works}

\paragraph{Cohort Explanation} Cohort explainability aims to strike a balance between local and global methods, which provide explanations based on a group of instances. It can be classified into inherent and post-hoc methods. Inherent cohort explanation, sometimes referred to as a local surrogate model, involves dividing the feature space and developing a local, interpretable model for each subspace. Examples of this approach include \citet{slim}, which constructs local-surrogate linear or generalized additive models by recursively partitioning the feature space.

Post-hoc cohort explanation methods partition the feature space and provide an explanation for each cluster. In general, such explainers run local explanation methods on a dataset and then cluster the samples based on the similarity between the local explanations. Such methods include \citet{glocal-shap}, which averages local SHAP values within predefined groups. Other methods employ techniques to automatically identify the cohorts, such as VINE, which clusters individual conditional expectation (ICE) via unsupervised clustering methods \citep{vine}. Similarly, REPID \citep{repid} and \citet{molnar2023model}, which use tree-based partitioning to generate the cohorts either based on ICE or conditional permutation feature importance. Additionally, GADGET extends previous methods by incorporating functional decomposition during partitioning, allowing for the consideration of more features \citep{gadget}. In contrast to the partitioning aspect, there are also works that focuses on how to aggregate local explanations, such as GALE \citep{gale} which uses homogeneity to re-weight LIME importance before aggregating for multi-class classification problems.

A common challenge for XAI in general is a lack of evaluation metrics to compare different methods, since the explanation template varies between methods. The most common metrics used to qualify cohort explanation is generalizability/coherence, the similarity between local explanations within cohorts. In this paper, we propose additional desiderata and evaluation metrics for cohort explanation. We also introduce a supervised-clustering-based approach that achieves better performance based on the metrics.

\paragraph{Subgroup Discovery} Subgroup discovery is a prominent area in data mining and machine learning that focuses on identifying interesting and interpretable subgroups within a dataset, which is valuable for uncovering hidden structures and actionable insights in various domains \citep{subgroup-survey}. Such works include \citet{sutton2020identifying} which uses rule-based partitioning based on the loss of surrogate models to identify interesting regions. Similarly, \citet{hedderich2022label} also use a rule-based algorithm on natural language models to recognize regions with significant response to certain variables. These methods can be thought of as a type of \textit{inherent} cohort explanation, though the identified clusters of interest may not cover the entire feature space. 

\paragraph{Explainable Clustering} Explainable clustering aims to enhance interpretability of clustering by providing understandable explanations for the derived clusters. For example, \citet{explainable-kmeans} modifies the results from $k$-means into decision trees to increase the interpretability of the clusters. \citet{xclusim} creates a visualization tool to compare and analyze clustering results. \citet{exp-cluster-optim} formulate a constraint optimization program to clustering to allow for the injection of domain knowledge. Though both domains utilize explainability and clustering, their goals are different. Cohort explanation aims to apply clustering to simplify the result of local explanations; constratively, explainable clustering's goal is to introduce explainability methods to existing clustering algorithms.

\section{Notation}
We will use the following notation throughout the paper. 

\begin{itemize}[noitemsep,topsep=0pt,parsep=0pt]
  \item $M$: target model, or the model to be explained;
  \item $x$: a particular sample;
  \item $X$ or $X_j$: a group of samples. We will use $X$ to denote the entire evaluation dataset and $X_j$ to denote the $j$-th cohort. $\{X_1, \dots, X_k\}$ is referred as a \textit{cohort definition}, which partitions the dataset $X$ into $k$ cohorts;
  \item $\omega$: a local explanation method. It is data driven and outputs local feature importance. Specifically, $\omega_M(X, x)$ is the result of evaluating $\omega$ in $x$, on the model $M$ and using $X$ as context samples. It should treat the target $M$ as a black-box, i.e., $\omega$ does not depend on the specific architecture of $M$;
  \item $e_j$: the explanation for cohort $j$;
  \item $a_x \in \{1, \dots, k\}$: the cohort assignment for sample $x$, representing which cohort $x$ belongs to.
\end{itemize}

\section{Desiderata}
\label{sec:desiderata}

\begin{table*}[tp]
\small
\centering
\begin{tabular}{c|ccc}
\toprule
 & \textbf{Global} & \textbf{Cohort} & \textbf{Local} \\ \hline
Generalizability & w.r.t the whole model & w.r.t. the cohorts & - \\ \hline
Conciseness & most concise & adjustable & least concise \\ \hline
Disjoint Cohorts& - & applicable & - \\ \hline
Cohort Locality & - & applicable & - \\ \hline
Stability & w.r.t. target model & \begin{tabular}[c]{@{}c@{}}w.r.t target model\\ and cohort definitions\end{tabular} & w.r.t. target model \\ \bottomrule
\end{tabular}
\caption{Desired properties of explanations at different scope.}
\label{tab:desiderata}
\end{table*}

Desired properties of XAI desiderata have been well analyzed in works such as \citet{survey2018}, \citet{survey2020} and \citet{fact-sheet}. In this section, however, we focus on properties that are unique to cohort explanations. Table~\ref{tab:desiderata} lists these properties and compares these desired properties of cohort explanations with their counterparts on a global or local scale.

\paragraph{Generalizability} Global explanation is prone to aggregation bias, where the average importance may not reflect the representative population in the dataset \citep{gadget}. Cohort explanation should aim to reduce such bias and to generate explanations that generalize well in each of the cohorts. If we apply a local method to all samples in a specific region, the explanation should be of low variance and close to the cohort explanation.

To evaluate generalizability, we use the average error w.r.t. local importance as the metric. Given a dataset $X$, a local importance method $\omega$, cohorts $\{X_j\}$ and the respective explanations $\{\bar{e}_j\}$, the loss is defined as 
\begin{align}
  \mathcal{L}_\mathrm{generalizability} = \frac{1}{|X|} \sum_{j=1}^k \sum_{x \in X_j} \|\omega_M(X_j, x) - \bar{e}_j\|_2^2
  \label{eq:generalizability}
\end{align}

\paragraph{Conciseness} One key goal of cohort explanation, especially cohort explanations, is to generate a concise explanation. It is evaluated using the number of cohorts and should be low without significantly sacrificing generalizability.

\paragraph{Disjoint Cohorts} Cohort explanation should yield meaningful cohort definitions. Although cohorts can be defined using samples, the regions represented by these assignments should divide the feature space into disjoint subspaces. Otherwise, it would be difficult to describe what a cohort represents. This adds an additional challenge in defining the cohorts in addition to the generalizability criteria. 

\paragraph{Cohort Locality} The explanation of a cohort should emphasize the decision-making \textit{local} or \textit{specific} to this region. That is, if two models have the same behavior within a given cohort $X$ (i.e., both models predict the labels for all samples in $X$), but they differ outside $X$, the cohort explanation for the two models in this fixed cohort should be the same.

However, this property does not hold in general for local explanation methods. Popular methods such as LIME \citep{lime} and KernelSHAP \citep{deepshap} require sampling, which may sample outside the cohort $X$. In this case, a good cohort explanation should try to preserve locality by stabilizing the explanation if a cohort is fixed.

To evaluate locality, given a fixed cohort $X_j$, we create alternative models $\tilde{M}_j$ that behave the same as the original model $M$ within $X_j$, but have the probability to be random otherwise. Specifically,
\begin{align}
  \tilde{M}_j(x) = \begin{cases}
    M(x), &x \in X_j, \\
    M(x), &x \not\in X_j, \text{w/ probability } 1-p, \\
    \text{random}, &x \not\in X_j, \text{w/ probability } p. \\
  \end{cases}
\end{align}
where the last case denotes a random class for classification tasks or a random value sampled from the distribution of labels for regression tasks. We then consider the average difference in explanation on cohort $X_j$ w.r.t. a probability $p$:
\begin{align}
  \mathcal{L}_\mathrm{locality}(p) = \mathbb{E}\left[ \sum_{j=1}^k \|e_j - \tilde{e}_j\|_2^2 \right].
  \label{eq:locality}
\end{align}
where $e_j$ is evaluated on target $M$ and $\tilde{e}_j$ is on $\tilde{M}_j$, and the expectation is over the randomness from $p$. An ideal cohort explanation should have a low locality loss and thus focus on localized information.

\paragraph{Stability} The explanation should be robust w.r.t. (1) the target model and (2) the cohort definitions. The cohort definitions should be stable and the importance should not drastically differ when the definitions are slightly changed.

To evaluate cohort stability, we tested the robustness of cohort definitions against the same input and then use the Adjusted Rand Index (ARI) \citep{ari} to compare the similarity between two assignments. To evaluate, we conduct $t$ run on the same dataset, and the metric is defined as 
\begin{align}
  \mathcal{S}_\mathrm{cohort~stability} = \frac{1}{t} \sum_{i=2}^t \mathrm{ARI}\big(\{a_x^1\}, \{a_x^i\}\big).
  \label{eq:stability_cohort}
\end{align}
where $a_x^i$ denotes the assignment computed on the $i$-th run.

To evaluate importance stability, we consider the change in explanation in a particular cohort if a new random sample is added to this cohort. Ideally, we want all samples to have localized, stable local importance scores. Thus, adding a sample to a cohort should not drastically affect its average importance. Therefore, we evaluate importance stability as 
\begin{align}
  \mathcal{L}_\mathrm{importance~stability} = \mathbb{E}_x\left[ \sum_{j=1}^k \|e_j - e_{X_j \cup \{x\}}\|_2^2 \right].
  \label{eq:stability_importance}
\end{align}
where $e_{X_j \cup \{x\}}$ is evaluated on the artificial cohort by appending sample $x$ to the cohort $X_j$. Note that Eqs.~\ref{eq:stability_cohort} and \ref{eq:stability_importance} measure two different aspects of cohort explanations. Cohort stability estimates the robustness on how the data are clustered, while importance stability concerns the robustness of explanations on predefined cohorts.

\section{Problem Definition}
\label{sec:def}

We emphasize on generate \textit{post-hoc} cohort explanation through existing \textit{local, data driven, feature importance} explanations. We formulate the cohort explanation as an optimization problem, with the goal of minimizing the generalizability loss in Eq.~\ref{eq:generalizability} while penalizing for having too many cohorts. Consider a local data-driven explanation method $\omega_M(X, x)$, where $M$ is the target model, $X$ is a context dataset and $x$ is a sample. The problem is defined as
\begin{align*}
    \min_{k, c_1, \dots, c_k} &
    \left(\frac{1}{|X|} \sum_{j=1}^k \sum_{x \in X_j} \|\omega_M(X_j, x) - e_j\|_2^2 \right) \\ 
  &+ \lambda \sqrt{\max\left(\frac{k - k^*}{|X|}, 0\right)}.
  \addtocounter{equation}{1}\tag{\theequation}
  \label{eq:obj}
\end{align*}

where $k$ is the number of cohorts, $c_j$ is the centroid that defines the cohort $X_j$:
\begin{align}
  X_j = \{x \in X ~|~ \|x - c_j\|_2 \le \|x - c_l\|_2, \forall l \ne j\}.
  \label{eq:cohort-def}
\end{align}

$e_j$ is explanation for the cohort $X_j$, defined as the average local importance across all samples in $X_j$:
\begin{align}
  e_j = \frac{1}{|X_j|} \omega_M(X_j, x).
  \label{eq:cohort-exp}
\end{align}

$\lambda$ and $k^*$ are hyperparameters and $k^*$ is the number of expected cohorts. $k^*$ should be chosen as the desired granularity of the explanations, similar to how the number of clusters is chosen in clustering algorithms. Note that setting $k^* = 1$ with a large enough $\lambda$ would generate global explanation, while setting it to the number of samples would be equivalent to local explanation. The first component in Eq.~\ref{eq:obj} represents the \textit{generalizability} criteria, while the second component corresponds to \textit{conciseness}. Choosing $\lambda$ would shift the balance between these two core desired properties of cohort explanations. Note that both the explanation in Eq.~\ref{eq:cohort-exp} and the objective function in Eq.~\ref{eq:obj} use the cohort $X_j$ instead of the whole dataset $X$ to compute local importance. This aims to satisfy the \textit{cohort locality} criteria by reducing information leakage from outside the cohorts. Additionally, since cohorts are defined using centroids as shown in Eq.~\ref{eq:cohort-def}, the partitions are guaranteed to be \textit{disjoint}.

\begin{algorithm*}[htp]
\small
\caption{Generalized cohort explanation conversion framework (CohEx)}
\label{alg:cohort}
\SetKwInOut{Input}{Input}
\SetKwInOut{Output}{Output}
\SetKwComment{Comment}{/* }{ */}
\Input{Number of iterations $n$,
Expected number of cohorts $k$,
Target model $M$, \newline
Dataset $X$, 
Data-driven local explanation method $\omega$, 
Supervised clustering algorithm $g$.
}
\Output{Cohort assignment $a_x \in [1, k], \forall x \in X$, 
Cohort explanations $\bar{w}_j, j \in [1, k]$.
}
  \For{i = 1, \dots, n}{
  randomly select $k$ centroids $c_1, \dots, c_k$ from $X$\;
  \For{$x \in X$}{
    $a_x^i \gets \arg\min_{1 \le a \le k} \|x - c_a\|_2$ \Comment*[r]{assign each sample to the closest centroids}
  }
  \Repeat{clustering loss does not decrease for t iterations}{
    \For{j = 1, \dots, k}{
      $X_j \gets \{x | a_x^i = j\}$ \Comment*[r]{The samples in this cohort}
      \For{$x \in X_j$}{
        $w^i_x \gets \omega_M(X_j, x)$ \Comment*[r]{Recompute explanations using samples only in the cohort}
      }
      $\bar{e}^i_j \gets (\sum_{x\in X_j} w^i_x) / |X_j| $ \Comment*[r]{Compute the average local explanations}
    }
    $k, \{a^i_x\} \gets g(k, X, \{w^i_x\})$ \Comment*[r]{Recluster using the new explanations}
  }
}
\textbf{return} the best $\{a^i_x\}$ and $\{\bar{e}^i_j\}$ with the lowest clustering loss
\end{algorithm*}

\section{Challenges}
\label{sec:challenge}

We now discuss two major challenges unique to cohort explanations, which our algorithm aims to overcome.

\begin{figure}[htp]
  \centering
  \includegraphics[width=0.6\linewidth]{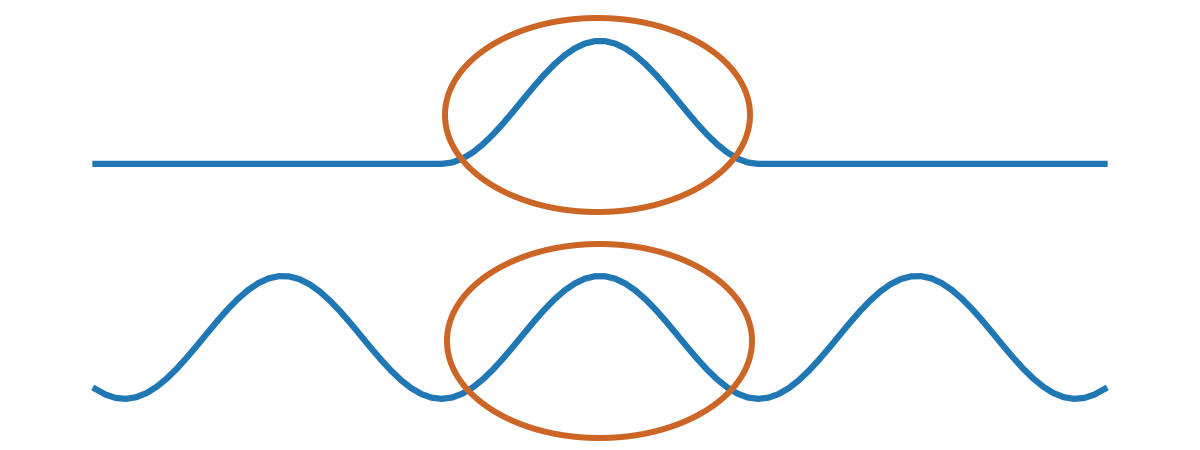}
  \caption{Two 1-D regression models with the same behavior on a specific cohort. The horizontal axis represents the feature, and the vertical axis represents the model output. The orange circle denotes the cohort.}
  \label{fig:locality-exp}
\end{figure}

\paragraph{Naive Average of Local Explanation Does Not Suffice } 
Given a particular cohort definition, the naive way to compute the cohort explanation is to average the local explanation of all members. However, this may fail the \textit{cohort locality} criteria discussed in Sec.~\ref{sec:desiderata}, since local explanations still contain information outside the cohort. For example, consider two regression models trained on the same tasks as shown in Fig.~\ref{fig:locality-exp}. On the cohort represented by the orange circle, the models have the same behavior. Naturally, we expect the cohort explanations to this cohort to be identical between the models as well. However, this is usually not the case based on the naive method. Data-driven local explanations either require a context dataset (e.g. SHAP) or need sampling (e.g. LIME) that would exceeds the boundary of the cohort. If we approach cohort explanation hierarchically - first compute all local importance, then apply clustering and/or transformations, which is true for most existing literature - the different behavior outside this cohort would leak into it, causing the cohort explanation to diverge. To address the issue without constraining the underlying local explainer, we propose an iterative framework for the conversion of local explanations that takes into account leakage and evaluates its locality based on Eq.~\ref{eq:locality}.

\paragraph{Unifying Cohorts and Explanations } Cohort explanations contain two distinct tasks: (1) creating definitions of the cohorts and (2) assigning an appropriate explanation on each of the cohorts. As discussed in Sec.~\ref{sec:desiderata}, all the generalizability, conciseness, and representativeness properties put constraints on the definition of the cohorts. 

Additionally, this is not a sequential task: one cannot first decide the cohort definitions and then compute their explanations, since the two tasks are interdependent. The cohort locality properties ensure that the explanation needs to be re-evaluated if the cohort definitions change, and the definitions depend on the explanations to achieve generalizability. 

\section{Approach}
\label{sec:approach}

To solve these two problems, we propose a generalized cohort explanation conversion framework (CohEx) to adapt data-driven, local, feature importance methods into cohort explanation. We allow modification to local importance to account for locality, and design a feedback loop between recomputing importance and reclustering to refine cohort definitions and their importances.

Algorithm~\ref{alg:cohort} shows the proposed method CohEx. The core idea is to provide only \textit{data-driven} local explanation methods with information within the cohort. In each iteration, we first randomly create cohort assignments through randomly sampled centroids at lines 2-4. Then we recompute all importances at line 9, using only samples within each cohort. Then we run a supervised clustering algorithm to create new cohort assignments using the new importances at line 13. This is repeated until the clustering loss does not decrease for a certain number of iterations. The whole process is then repeated $n$ times and the best result is reported.

To generate the cohort definitions, we choose supervised clustering.  Compared to traditional clustering, it assumes that data are labeled and uses both features $X$ and labels $y$ to group data. Each data sample is clustered into the closest centroids in the \textit{feature space}, while the goodness of an assignment is based on the \textit{labels}, or in our case, feature importance (Eick et al., 2004). This ensures that the cohort definitions are always closed and well defined, while constraining local explanations within the cohort to be similar for the generalizability criteria. Specifically, in our evaluations, we use Single Representative Insertion/Deletion Steepest Decent Hill Climbing with Randomized Restart (SRIDHCR) as the supervised learning method, a greedy algorithm that attempts to solve the problem by randomly adding or removing centroids \citep{supervised-clustering}. Note that standard, \textit{unsupervised} clustering algorithms such as HDBSCAN does not apply here since we need a \textit{supervised} mechanism to cluster both the feature values and the importance scores simultaneously. 

For the convergence of the algorithm, since the cohort assignments depend on the base local explanation methods, the iterative loop between recomputing importance and reclustering is not guaranteed to converge. We resolved to use clustering loss as a stopping criterion. The algorithm terminates if the loss does not decrease for a set number of iterations, and we repeat this process $n$ times with different initialization. Thus, the algorithm is guaranteed to terminate, though the convergence speed may depend on the specifics of the chosen base method. The complexity analysis of CohEx and discussion related to non-data-driven methods can be found in Appendix~A and B.

\begin{figure*}[ht]
  \centering
  \begin{subfigure}[t]{0.15\textwidth}
    \centering
    \includegraphics[width=\textwidth]{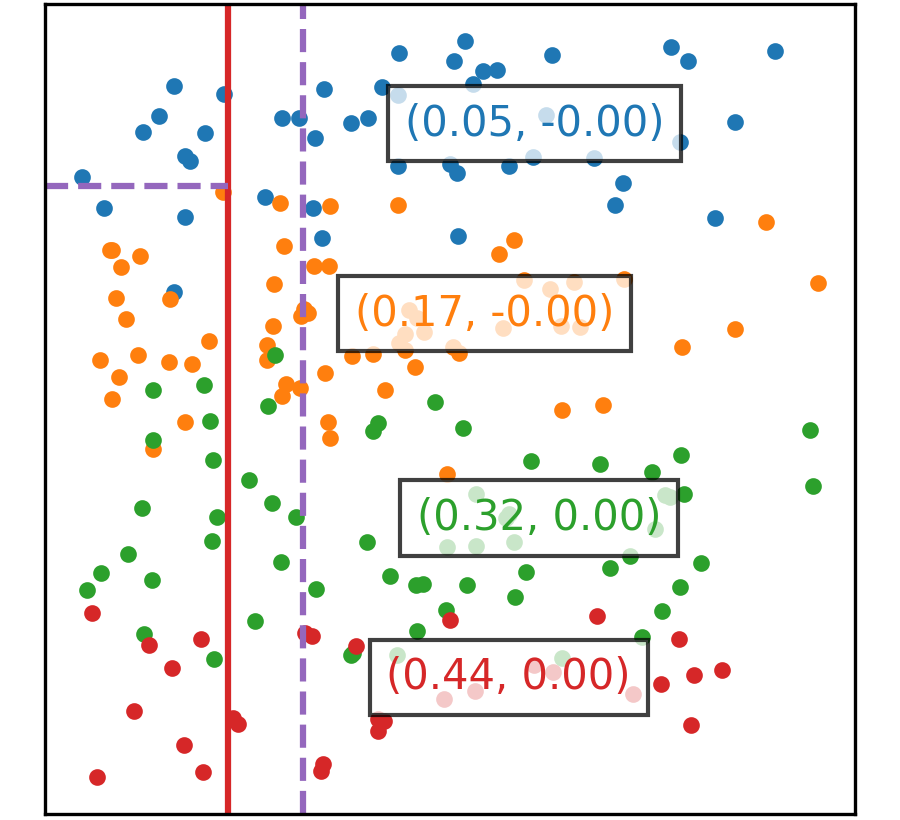}
    \caption{VINE.}
  \end{subfigure}
  \begin{subfigure}[t]{0.15\textwidth}
    \centering
    \includegraphics[width=\textwidth]{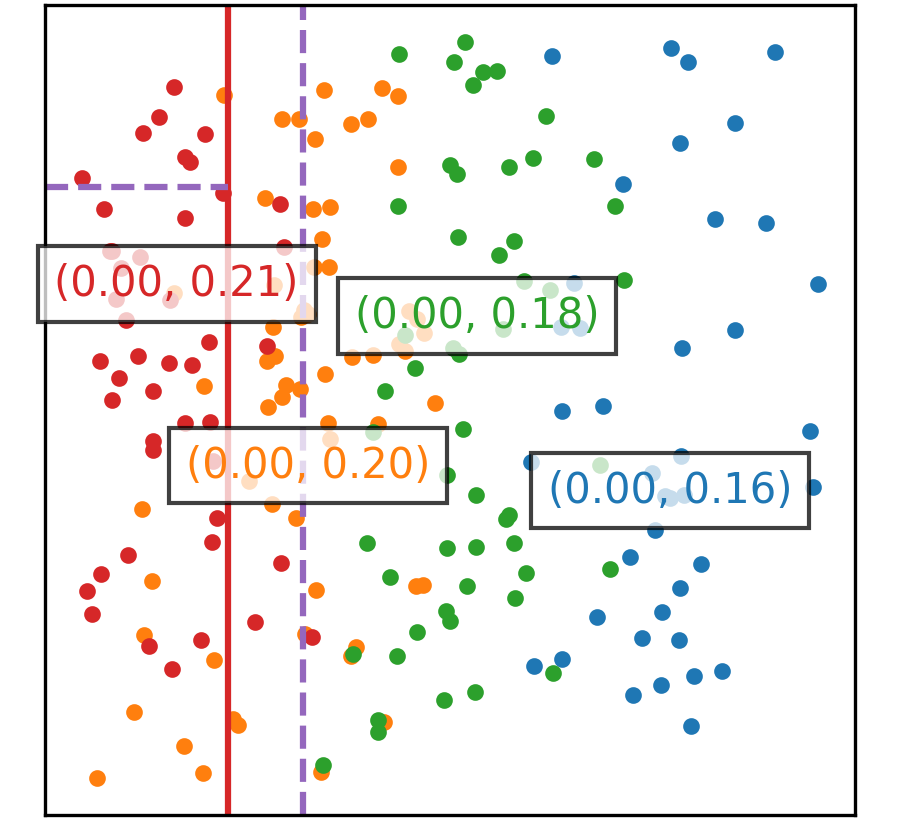}
    \caption{VINE+GALE.}
  \end{subfigure}
  \begin{subfigure}[t]{0.15\textwidth}
    \centering
    \includegraphics[width=\textwidth]{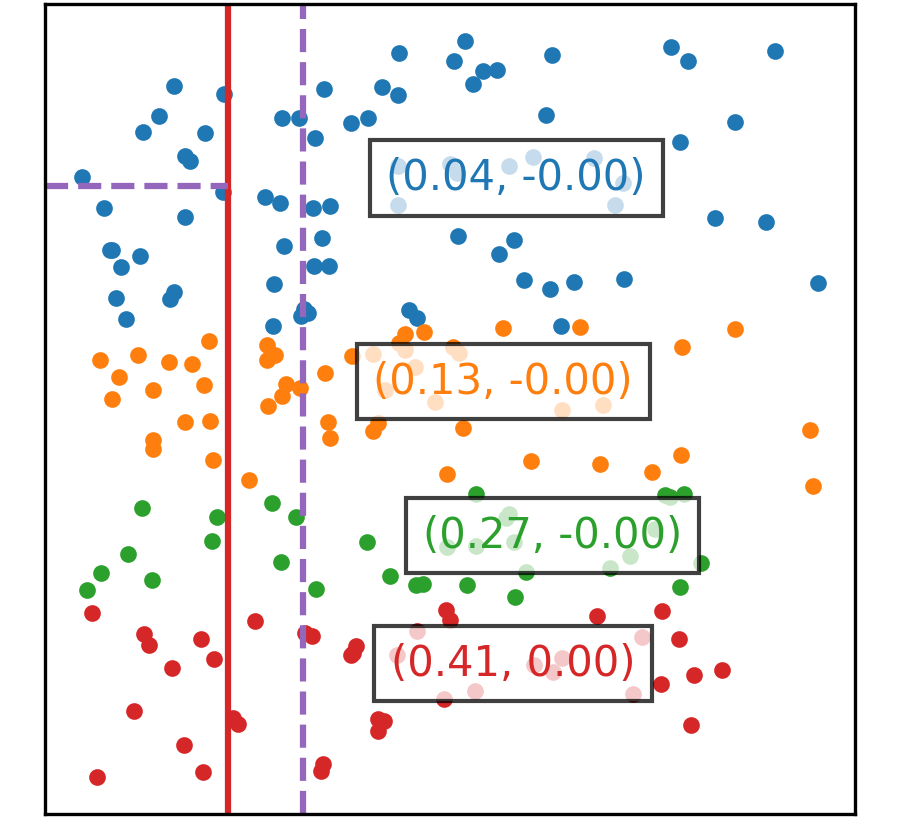}
    \caption{REPID.}
  \end{subfigure}
  \begin{subfigure}[t]{0.15\textwidth}
    \centering
    \includegraphics[width=\textwidth]{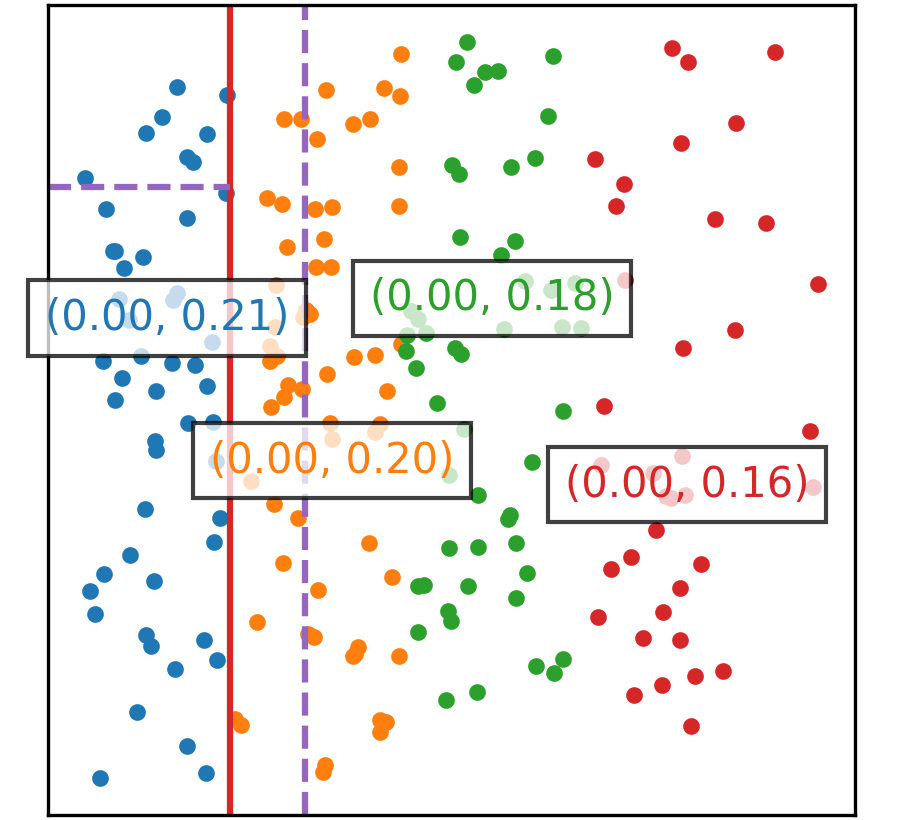}
    \caption{REPID+GALE.}
  \end{subfigure}
  \begin{subfigure}[t]{0.15\textwidth}
    \centering
    \includegraphics[width=\textwidth]{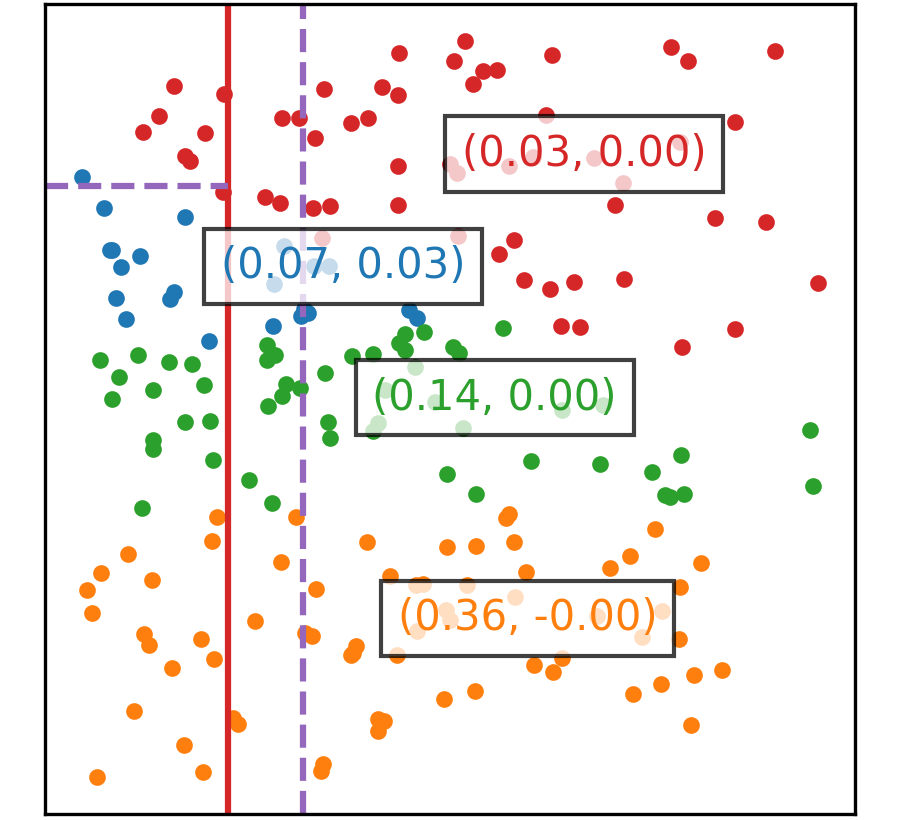}
    \caption{Hierarchical.}
  \end{subfigure}
  \begin{subfigure}[t]{0.15\textwidth}
    \centering
    \includegraphics[width=\textwidth]{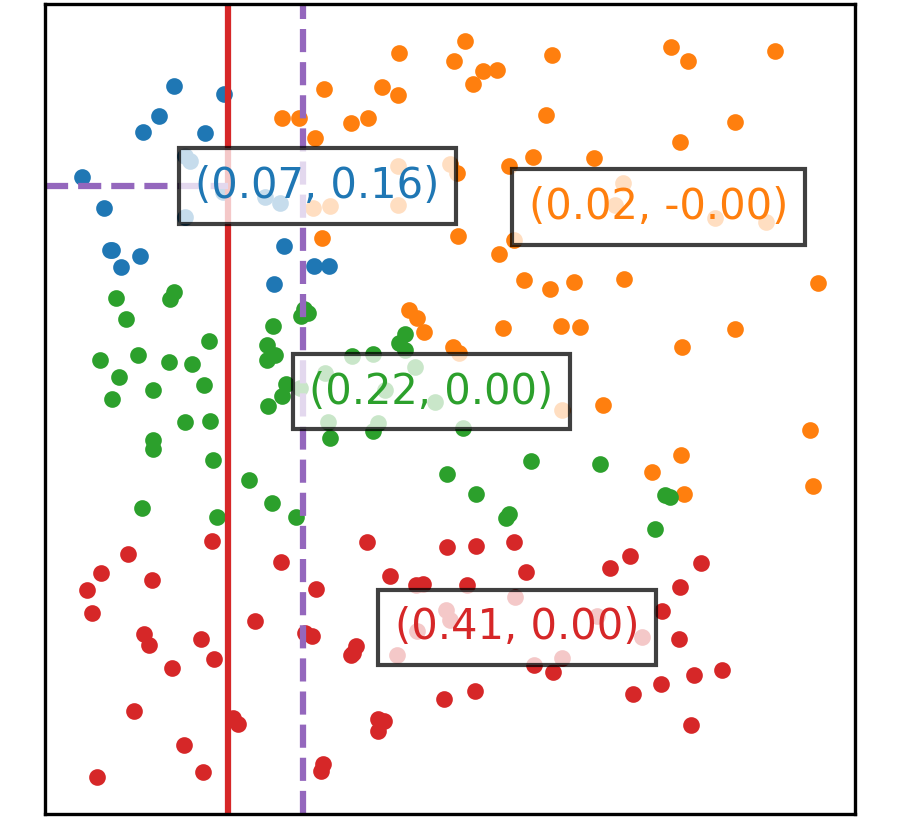}
    \caption{CohEx.}
    \label{fig:motivation-result-cohex}
  \end{subfigure}
  
  \caption{Running the six cohort explanation algorithms on the motivational example in Fig.~\ref{fig:motivation}. Each color represents a different cohort. The labels in the figures denotes cohort explanation $(e_\mathrm{age}, e_\mathrm{family})$ of the cohort of the corresponding color.}
  \label{fig:motivation-result}
\end{figure*}

\begin{table*}[h]
\small
\centering
\begin{tabular}{@{}cccccc@{}}
\toprule
 & Clustering Penalty ($\downarrow$) & Closed Cohort & Cohort Locality ($\downarrow$) & Cohort Stability ($\uparrow$) & Importance Stability ($\downarrow$) \\ \midrule
\multicolumn{6}{c}{\textbf{Synthetic Patient Classification (LIME)}} \\ \midrule
\multicolumn{1}{c|}{VINE} & $0.0011 \pm 0.0001$ & $\times$ & $0.0273 \pm 0.0076$ & $42.33\% \pm 0.78\%$ & $0.0131 \pm 0.0039$ \\
\multicolumn{1}{c|}{VINE+GALE} & $\mathbf{0.0006 \pm 0.0000}$ & $\times$ & $0.1011 \pm 0.0046$ & $42.96\% \pm 5.38\%$ & $0.0481 \pm 0.0070$ \\
\multicolumn{1}{c|}{REPID} & $0.0012 \pm 0.0000$ & $\checkmark$ & $0.0301 \pm 0.0021$ & $47.37\% \pm 0.56\%$ & $0.0169 \pm 0.0015$ \\
\multicolumn{1}{c|}{REPID+GALE} & $\mathbf{0.0005 \pm 0.0000}$ & $\checkmark$ & $0.0978 \pm 0.0039$ & $\mathbf{53.52\% \pm 9.92\%}$ & $0.0415 \pm 0.0048$ \\
\multicolumn{1}{c|}{Hierarchical} & $0.0018 \pm 0.0003$ & $\checkmark$ & $0.0157 \pm 0.0050$ & $36.74\% \pm 9.57\%$ & $0.0167 \pm 0.0019$ \\
\multicolumn{1}{c|}{CohEx} & $\mathbf{0.0006 \pm 0.0003}$ & $\checkmark$ & $\mathbf{0.0052 \pm 0.0088}$ & $40.76\% \pm 10.34\%$ & $\mathbf{0.0027 \pm 0.0033}$ \\ \midrule
\multicolumn{6}{c}{\textbf{Bike Sharing (SHAP)}} \\ \midrule
\multicolumn{1}{c|}{VINE} & $0.6641 \pm 0.0198$ & $\times$ & $3.5983 \pm 0.3986$ & $79.89\% \pm 17.87\%$ & $3.0773 \pm 0.5905$ \\
\multicolumn{1}{c|}{REPID} & $1.1633 \pm 0.0000$ & $\checkmark$ & $2.5280 \pm 0.0091$ & $\mathbf{100.00\% \pm 0.00\%}$ & $2.1123 \pm 0.0046$ \\
\multicolumn{1}{c|}{Hierarchical} & $0.4796 \pm 0.1173$ & $\checkmark$ & $3.1891 \pm 0.0187$ & $39.17\% \pm 10.11\%$ & $2.5951 \pm 0.0061$ \\
\multicolumn{1}{c|}{CohEx} & $\mathbf{0.3196 \pm 0.1018}$ & $\checkmark$ & $\mathbf{0.5034 \pm 0.2214}$ & $43.68\% \pm 14.23\%$ & $\mathbf{0.1022 \pm 0.0886}$ \\ \midrule
\multicolumn{6}{c}{\textbf{MNIST (DeepLIFT)}} \\ \midrule
\multicolumn{1}{c|}{VINE} & $1.5450 \pm 0.1698$ & $\times$ & $0.0528 \pm 0.0253$ & $80.98\% \pm 22.07\%$ & $4.1130 \pm 0.932$ \\
\multicolumn{1}{c|}{VINE+GALE} & $1.4729 \pm 0.0918$ & $\times$ & $0.1417 \pm 0.0253$ & $81.03\% \pm 19.28\%$ & $2.1209 \pm 0.634$ \\
\multicolumn{1}{c|}{REPID} & $\mathbf{1.3462 \pm 0.0000}$ & $\checkmark$ & $0.0531 \pm 0.0198$ & $\mathbf{100.00\% \pm 0.00\%}$ & $1.4889 \pm 0.0060$ \\
\multicolumn{1}{c|}{REPID+GALE} & $\mathbf{1.3290 \pm 0.0000}$ & $\checkmark$ & $0.0999 \pm 0.0203$ & $\mathbf{100.00\% \pm 0.00\%}$ & $0.4139 \pm 0.0046$ \\
\multicolumn{1}{c|}{Hierarchical} & $1.9371 \pm 0.3825$ & $\checkmark$ & $0.0127 \pm 0.0047$ & $31.50\% \pm 32.08\%$ & $0.0956 \pm 0.0475$ \\
\multicolumn{1}{c|}{CohEx} & $\mathbf{1.3112 \pm 0.3037}$ & $\checkmark$ & $\mathbf{0.0010 \pm 0.0019}$ & $33.22\% \pm 29.13\%$ & $\mathbf{0.0210 \pm 0.0419}$  \\ \bottomrule
\end{tabular}
\caption{Evaluating the cohort explanation algorithms in the motivational example shown in Fig.~\ref{fig:motivation} and the bike sharing dataset. The ($\downarrow$) symbol after the criteria represents that the lower the better, and ($\uparrow$) represents the higher the better.}
\label{tab:motivation-result}
\end{table*}

\section{Evaluation}
\label{sec:evaluation}

To evaluate the proposed method, we experiment on three scenarios: (1) the patient classification problem in Sect.~\ref{sec:intro}, a small scale example to demonstrate the importance of using supervised clustering and revising importance; (2) bike sharing, a classical ML regression problem to predict the hourly number of rented bikes \citep{bikesharing}, and (3) MNIST digit classification \citep{mnist}, a vision-based task with deep learning to mimic more realistic and complex scenarios. For baselines, we compare with the following methods:
\begin{itemize}[noitemsep,topsep=0pt,parsep=0pt]
  \item \textbf{VINE}, which applies supervised clustering on the local importances scores \citep{vine};
  \item \textbf{REPID}, a tree-based partitioning method \citep{repid}. For the sake of comparison, we modify the algorithm to use a local explainer that is consistent with other method, instead of the original partial dependency explainer; \footnote{Python's SHAP package implements an auto-cohort feature. Though no reference is provided, we believe that it is equivalent to REPID.}
  \item \textbf{GALE}, a homogeneity-based re-weighting mechanism to improve the quality of importance aggregation in classification problems \citep{gale}. GALE is not a cohort explainer by definition, but we feed the GALE importance into VINE and REPID to assess the quality of the reweighed importance. Note that GALE is not applicable to regression problems;
  \item \textbf{Hierarchical cohort explanation}: first compute the local importance, then run supervised clustering with SRIDHCR once. The difference between this method and the proposed CohEx is that it does not iteratively recompute the local importance scores. 
\end{itemize}

\subsection{Synthetic Patient Classification}
\label{sec:eval-motivation}

\begin{figure*}[h]
  \centering
  \includegraphics[width=0.8\linewidth]{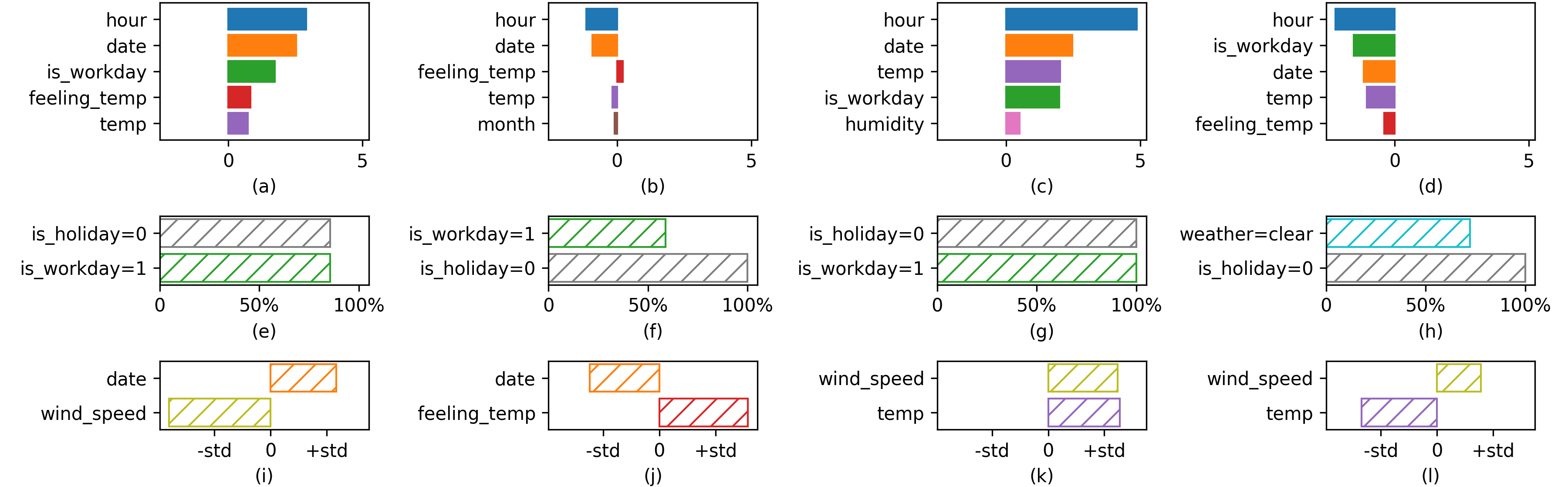}
  \caption{Running CohEx on the bike sharing dataset. Each column represents a cohort. (a-d) Cohort importance of the top five features in each cohort; (e-h) the top two most homogeneous categorical features in each cohort, measured in homogeneity; (i-l) the top two most abnormal continuous features, measured in the centroid difference w.r.t. dataset mean in standard deviations.}
  \label{fig:bike-result-cohex}
\end{figure*}

We continue to use the motivating example in Sect.~\ref{sec:intro}. For this evaluation, the expected number of cohorts $k^*$ is set to 4, and the maximum depth of the tree in REPID is set to 2 so that the number of cohorts is consistent for all methods. We use LIME as the base method, specifically since it does not satisfy the cohort locality criteria and allows us to compare the effect of different algorithms on locality. Fig.~\ref{fig:motivation-result} and Table~\ref{tab:motivation-result} show the result of running the algorithms on this example. The cohort partitioning and explanations are shown in Fig.~\ref{fig:motivation-result}. Additionally, the evaluation against different $k^*$ can be found in Appendix~C.1.

\subsection{Bike Sharing}
\label{sec:eval-bike}

This experiment focuses on a realistic machine learning regression problem. The bike sharing task asks to predict the hourly number of rented bikes between years 2011 and 2012 in the Capital bikeshare system based on seasonal and weather information. For the target model, we train an XGBoost model \citep{xgboost} on this dataset and achieve a $R^2$ score of 0.9453. We continue to use $k^* = 4$ or a depth of two for a similar number of cohorts and use SHAP as the local explainer.

Fig.~\ref{fig:bike-result-cohex} shows the result of running our proposed algorithm on the XGBoost model, including both the explanations and the characteristics to each cohort. Comparison with other methods can be found in the Appendix.~C.2. Our algorithm can detect clusters with distinct characteristics and importance signals, while maintaining a coherent explanation within each cohort. 

\subsection{Digit Classification}
We evaluate based on MNIST \citep{mnist}. Note that the MNIST inputs are cropped, centered, and relatively simple, so measuring the distance between importance heatmaps is meaningful. We built the target model as a neural network with two convolutional layers and two fully-connected layers and achieved an accuracy of 93.93\% on the test dataset. More details on the model and a visualization of the cohort explanation can be found in Appendix.~C.3.

\subsection{Evaluation Metrics and Analysis}

\paragraph{Generalizability \& Conciseness} CohEx achieves the best performance based on the objective function in Eq.~\ref{eq:obj}. Note that all methods yield $k=4$, so the conciseness loss for all is zero. VINE yields a low generalizability loss since its clustering criterion focuses only on feature importance, while all other methods need to consider the distance in the feature space. However, CohEx is able to exceed VINE since they use different local importance scores $w_x$. Through the recomputing of importance, CohEx edits the importance by putting more emphasis on local structures, allowing the importance scores in each cohort to have lower variance. Introducing GALE re-weighting improves the performance of VINE and REPID since it also recomputes importance, but it does not perform well on low-dimensional features as shown in Fig.~\ref{fig:motivation-result} since the importance of the feature with the lowest homogeneity will always be set to zero.

\paragraph{Cohort Locality} CohEx yields a significantly lower locality loss compared to the other methods according to Eq.~\ref{eq:locality}, which is as expected since CohEx introduces an iterative refinement process specifically for locality. To further justify the approach, in the patient classification problem, CohEx identifies a clustering that mostly match with the three clusters in our ideal explanation proposed in Sect.~\ref{sec:intro}. In particular, CohEx is the only algorithm that identifies a region (the blue cohort in Fig.~\ref{fig:motivation-result-cohex}) in which family history is the more important feature. Note that for almost all samples, the static LIME importance of age is higher than family history similar to the result in Fig.~\ref{fig:motivation-exp}. Therefore, since other methods are based on a static set of local importance scores, they cannot find any regions with a meaningful family history importance. We further justify that such a region should exist, since if we only consider said region, the horizontal decision boundary is the dominant decision boundary. This shows that CohEx's explanation is more \textit{local} to its cohorts. We also note that applying GALE would harm locality, since the re-weighting process smoothes the importance over the whole dataset, specifically contradicts the goal of locality.

\paragraph{Disjoint Cohorts} Only VINE fails this property since it clusters solely based on importance and it's very likely that some distant samples happen to have similar explanations and thus are partitioned together. All other method guarantees that the cohorts can be separated into disjoint areas by hyperplanes.

\paragraph{Cohort Stability} In all scenarios, REPID achieves higher ARI scores based on Eq.~\ref{eq:stability_cohort}, which shows that our supervised clustering-based approaches introduce variance. We consider this variance to be inherent to the methods, as both the recomputing process and SRIDHCR are trial-and-error by nature. The question of how to improve the stability remains a challenge for future work. 

\paragraph{Importance Stability} It measures the change in cohort explanations, if a random sample is added to the cohort. For fair comparisons, we allow the baseline methods to recompute the importances similar to CohEx after adding the sample, which greatly improve their stabilities. CohEx still exceeds the baselines. This demonstrates that even though the partitions of CohEx is not as robust based on the previous \textit{cohort stability} metric, on a fixed cohort, CohEx's explanation is much less gullible to changes or noises in the dataset.

\section{Conclusion}
\label{sec:conclusion}

In this paper, we identify a less explored piece in the XAI literature, i.e., cohort explanation. We illustrate the desired properties and challenges unique to the cohort explanation. We proposed a unified framework CohEx to generate such explanation by transferring existing local explanation methods while preserving generalizability and locality. It showed an increase in most metrics compared to baseline methods. 

The current limitations to the algorithm include (1) the stability of cohort definitions; (2) the high time complexity in more complex scenarios, such as vision tasks; (3) the limits of the framework to local data-driven explainers. For future directions, alternatives can be considered, such as using non-data-driven, non-local, or non-post-hoc explanation methods to generate cohort explanation, increasing generalizability to more complex tasks. Additionally, the meaningfulness of the cohort definitions can be potentially increased through enforcing stricter constraints on the definitions, such as only partitioning the samples using tags, rules, or hyperplanes.

\appendix
\section{Complexity Analysis of CohEx}
\label{app:complexity}

To ensure that the proposed CohEx algorithm terminates, it terminates when the supervised clustering loss does not decrease for a set number of iterations. Under this termination scheme, the complexity of CohEx is then $O(nmkT)$, where $n$ is the number of trials, $m$ is the average number of iterations within each trial, $k$ is the average number of cohorts throughout the clustering process, and $t$ is the average run time of applying the base local algorithm on each of the cohorts. Note that both $m$ and $T$ are directly affected by the choice of the local explainer. $T$ is usually related to the number of features or the size of the cohorts, while the stability of the algorithm may affect $m$. It is especially costly when the local explainer requires sampling, such as KernelSHAP. We suggest to tune the hyperparameter of the base method, such as the sampling size, for such algorithms.

\section{CohEx w/ Non-Data-Driven Methods}
Although we focus on adapting data-driven methods, CohEx can be modified if any of these premises cannot be met. If the base method is not data driven, for example, saliency map \citep{saliency}, vanilla gradient, SmoothGrad \citep{smoothgrad} or Grad$\times$Input \citep{gradxinput}, the explanations will not change during recomputing since $w$ does not depend on the input dataset $X$. In this case, similar to cohort explanations, we only need to run one pass of lines 6-13 in Algorithm 1. The algorithm fallbacks to a hierarchical scheme that first computes the importances, then uses supervised clustering to create the cohort definitions. However, as discussed previously, without the iterative loop, the result may not satisfy the cohort locality property.

\section{Additional simulation results}

\subsection{Synthetic Patient Classification}
\label{app:motivation}

In this section, we provide additional details of the synthetic patient classification experiment and additional results using alternative base local explanation methods. 

\subsubsection{Data and model generation}

The dataset in Fig.~\ref{fig:motivation-data} is generated as follows: let $A_i$ denote the age of the sample $x_i$ and $H_i$ denote its value of family history. Then
\begin{align}
  A_i &\sim 10B_i + Z_i \\
  B_i &\sim \mathrm{Discrete}(0, \dots, 9; & \nonumber \\
  &0.14, 0.15, 0.14, 0.15, 0.15, & \nonumber \\
  &0.11, 0.07, 0.06, 0.02, 0.01)& \\
  Z_i &\sim \mathrm{Unif}(0, 10)  \\
  H_I &\sim \mathrm{Unif}(0, 1)
\end{align}

We assume a optimal decision boundary of 
\begin{align}
  f(A, H) = \big(4(A / 100)^2 + (0.75H)^2\big) = 0.4
\end{align}

And the label is generated as 
\begin{align}
  Y_i|f(A_i, H_i) < 0.4 &= \begin{cases}
    0, &\mathrm{with~prob.~} 0.8, \\
    1, &\mathrm{with~prob.~} 0.2
  \end{cases} \\
  Y_i|f(A_i, H_i) \ge 0.4 &= \begin{cases}
    0, &\mathrm{with~prob.~} 0.2, \\
    1, &\mathrm{with~prob.~} 0.8
  \end{cases}
\end{align}

\subsubsection{Generalizability vs. conciseness}

\begin{figure}[htp]
  \centering
  \includegraphics[width=0.4\textwidth]{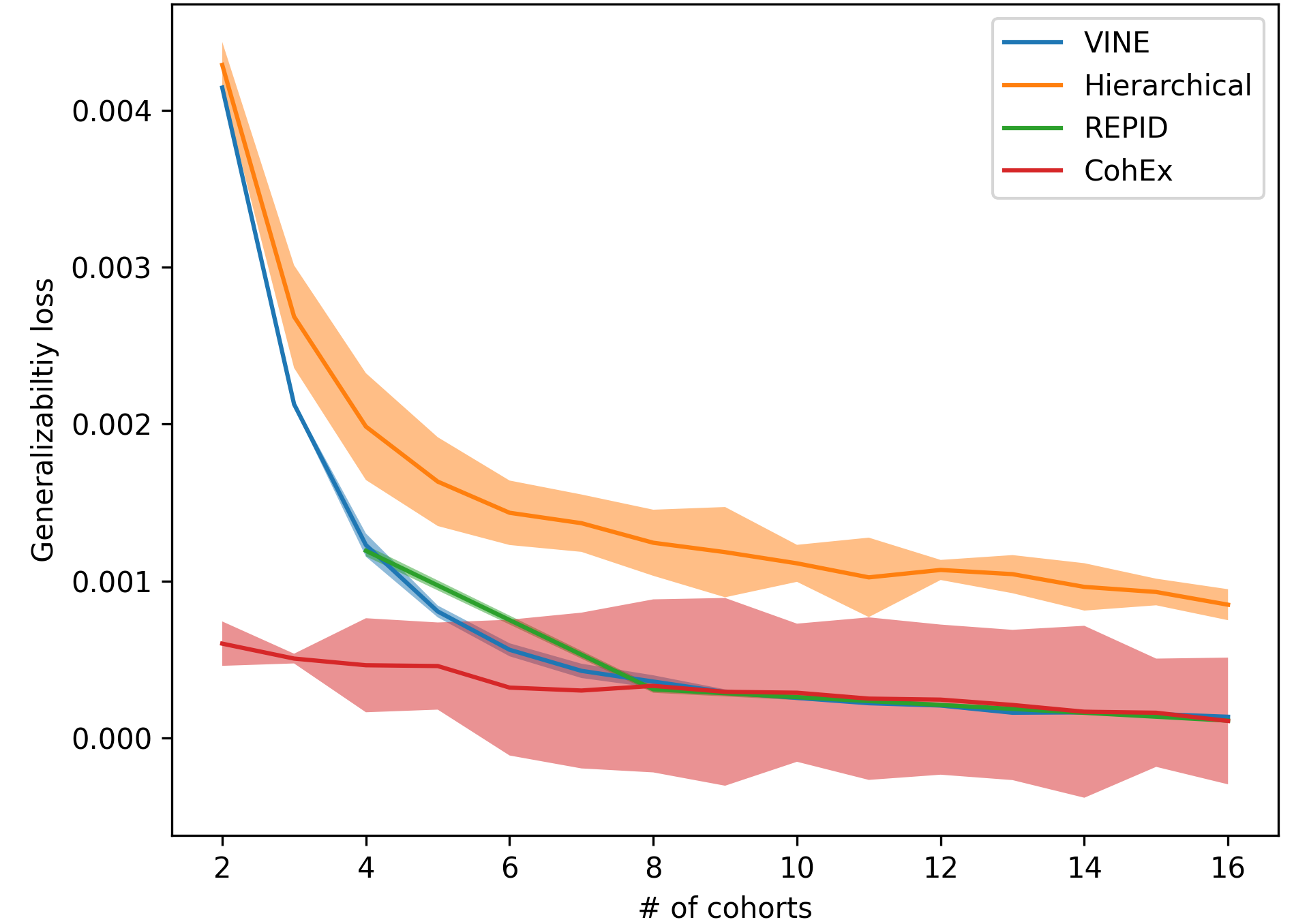}
  \caption{Generalizabilty loss vs. number of cohorts for the four cohort explanation algorithms on the patient classification dataset. The solid lines represent the average clustering loss among ten trials, and the shaded areas are within one standard deviation. Note that REPID is only evaluated at 4, 9 and 16 cohorts, with a respective max tree depth of 2, 3 and 4.}
  \label{fig:g-vs-c}
\end{figure}

Fig.~\ref{fig:g-vs-c} shows the change in generalizabilty loss for the four algorithms on this example. For the two algorithms based on supervised clustering, the final number of cohorts $k$ is the same as the expected number of cohorts $k^*$, so the loss of clustering is exactly the same as the generalizability loss component in Eq.~\ref{eq:generalizability}. The rankings of the four algorithms remain the same regardless of the number of cohorts.

We can interpret this figure as the balance between generalizability vs. conciseness. When the number of cohorts is small, the result approaches a global explanation with a concise but less quality explanation; when the number of cohorts becomes larger, the result approaches local explanation, whose explanation is more accurate but less concise. The selection of the desired number of cohorts $k^*$ represents the desired granularity of the cohort explanation. 

On average, the generazability loss of CohEx is able to converge much faster than the other algorithms due to its iterative local importance recomputing process, although the combination of supervised clustering and iteration introduces variance to the system, especially at high $k^*$, which is related to the cohort stability metric. Addressing the robustness of the framework remains one of the main goals for our future direction.

\subsubsection{Alternative base method}
In Sect.~8, we evaluated CohEx with other baseline methods while using LIME as the base local explainer. In this section, we provide additional evaluations if we use SHAP as the base method. 

\begin{figure}[ht]
  \centering
  \includegraphics[width=0.75\linewidth]{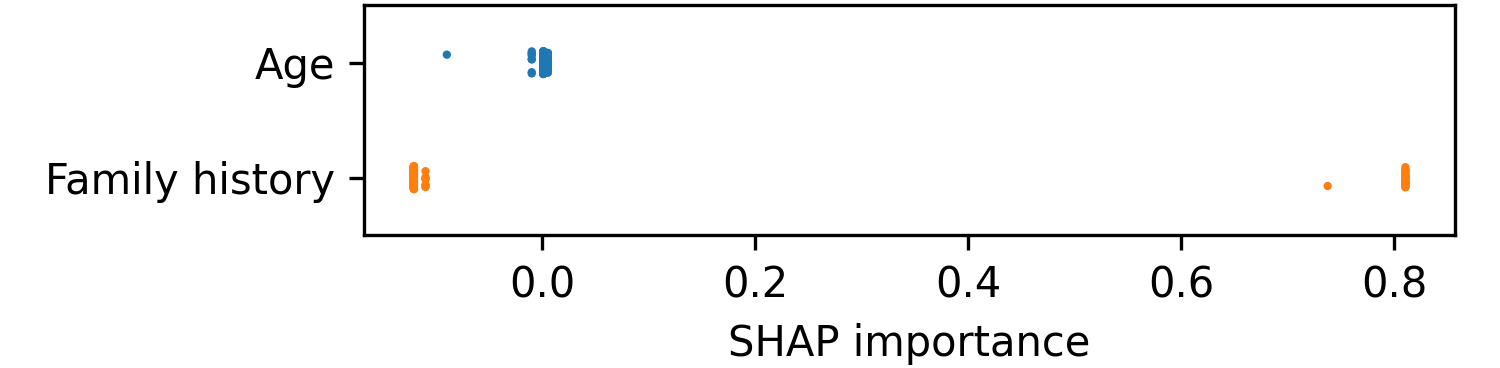}
  \vspace{-1em}
  \caption{The distribution of SHAP importance (using all data samples as the context) on the motivational medical classification example.}
  \label{fig:motivation-base-importance-shap}
\end{figure}

Table~\ref{tab:motivation-shap} shows the evaluation of the proposed criterion using SHAP. The comparison between the methods is mostly identical, though CohEx is now performing worse based SHAP. This is because the distribution of the base SHAP values is much less similar among the samples, which can be seen from the distribution in Fig.~\ref{fig:motivation-base-importance-shap}. VINE and REPID are able to partition the samples almost perfectly, and thus achieve a near-zero penalty. The drawback is that the reported importance scores are less informative: a few cohorts would yield the same importance. CohEx on the other hand recomputes the importance scores based on local cohorts, and thus reports a more nuanced importance, though at the cost of a slightly worse clustering penalty.

\begin{table*}[htp]
\small
\centering
\begin{tabular}{@{}cccccc@{}}
\toprule
 & Clustering Penalty ($\downarrow$) & Disjoint Cohort & Cohort Locality ($\downarrow$) & Cohort Stability ($\uparrow$) & Importance Stability ($\downarrow$) \\ \midrule
\multicolumn{6}{c}{\textbf{Synthetic Patient Classification (SHAP)}} \\ \midrule
\multicolumn{1}{c|}{VINE} & $\mathbf{0.0000 \pm 0.0000}$ & $\times$ & $0.5504 \pm 0.0007$ & $\mathbf{1.0000 \pm 0.0000}$ & $0.6175 \pm 0.0000$ \\
\multicolumn{1}{c|}{REPID} & $\mathbf{0.0000 \pm 0.0000}$ & $\checkmark$ & $0.5503 \pm 0.0005$ & $\mathbf{1.0000 \pm 0.0000}$ & $0.6175 \pm 0.0000$ \\
\multicolumn{1}{c|}{Hierarchical} & $0.0350 \pm 0.0151$ & $\checkmark$ & $0.2406 \pm 0.1084$ & $0.3564 \pm 0.0708$ & $0.2915 \pm 0.0579$ \\
\multicolumn{1}{c|}{CohEx} & $0.0091 \pm 0.0028$ & $\checkmark$ & $\mathbf{0.1303 \pm 0.0737}$ & $0.3889 \pm 0.1280$ & $\mathbf{0.0650 \pm 0.0999}$  \\ \bottomrule
\end{tabular}
\caption{Evaluating the cohort explanation algorithms in the patient classification task, using vanilla SHAP as the base local explanation method. The ($\downarrow$) symbol after the criteria represents that the lower the better, and ($\uparrow$) represents the higher the better.}
\label{tab:motivation-shap}
\end{table*}

\subsection{Bike sharing}
\label{app:bike}

In this section, we provide additional details about the bike sharing experiment.

\subsubsection{Result from other methods}

\begin{figure*}[tp]
  \centering
  \begin{subfigure}{\textwidth}
    \centering
    \includegraphics[width=\linewidth]{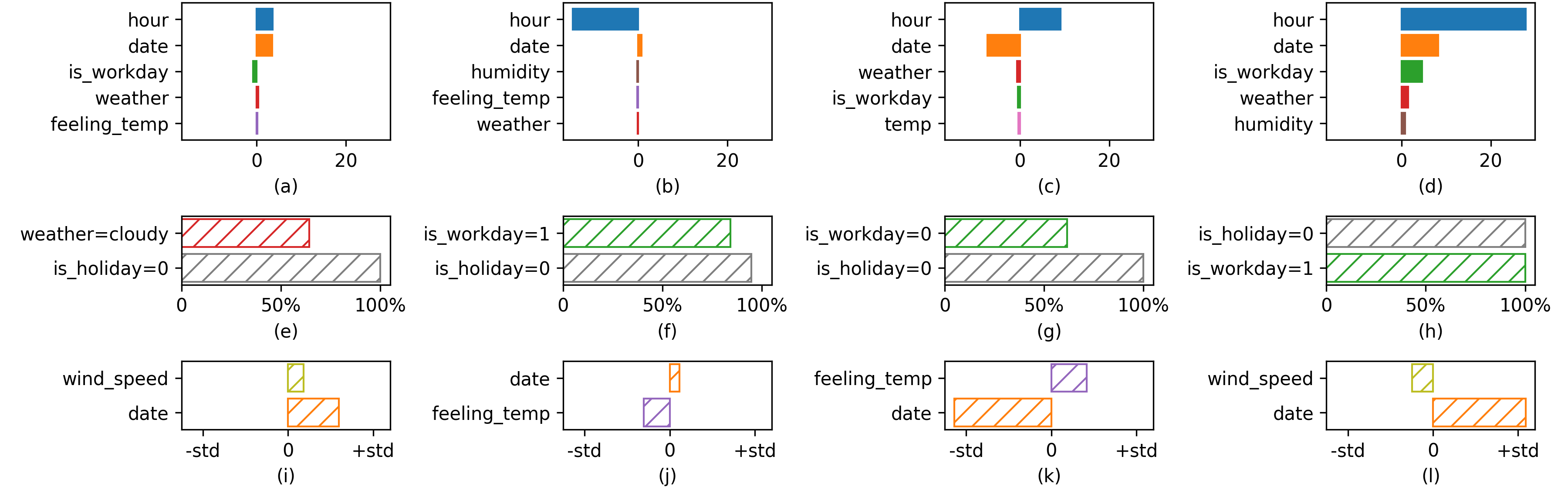}
    \vspace{-2em}
    \caption{VINE.}
  \end{subfigure}
  \begin{subfigure}{\textwidth}
    \centering
    \includegraphics[width=\linewidth]{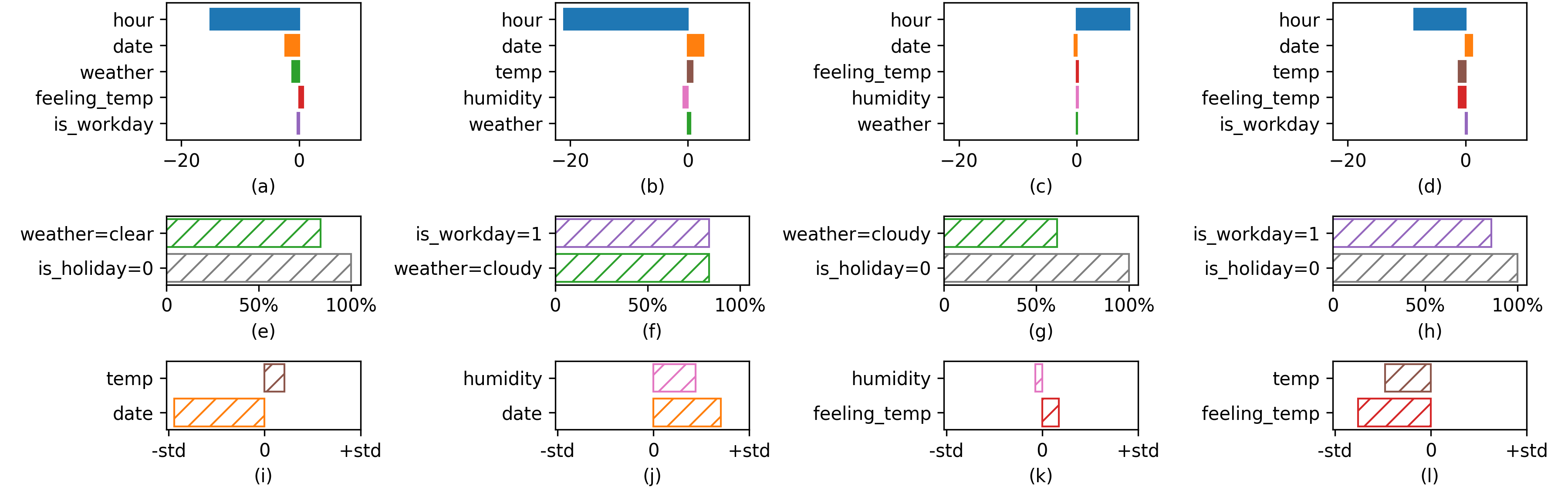}
    \vspace{-2em}
    \caption{REPID.}
  \end{subfigure}
  \begin{subfigure}{\textwidth}
    \centering
    \includegraphics[width=\linewidth]{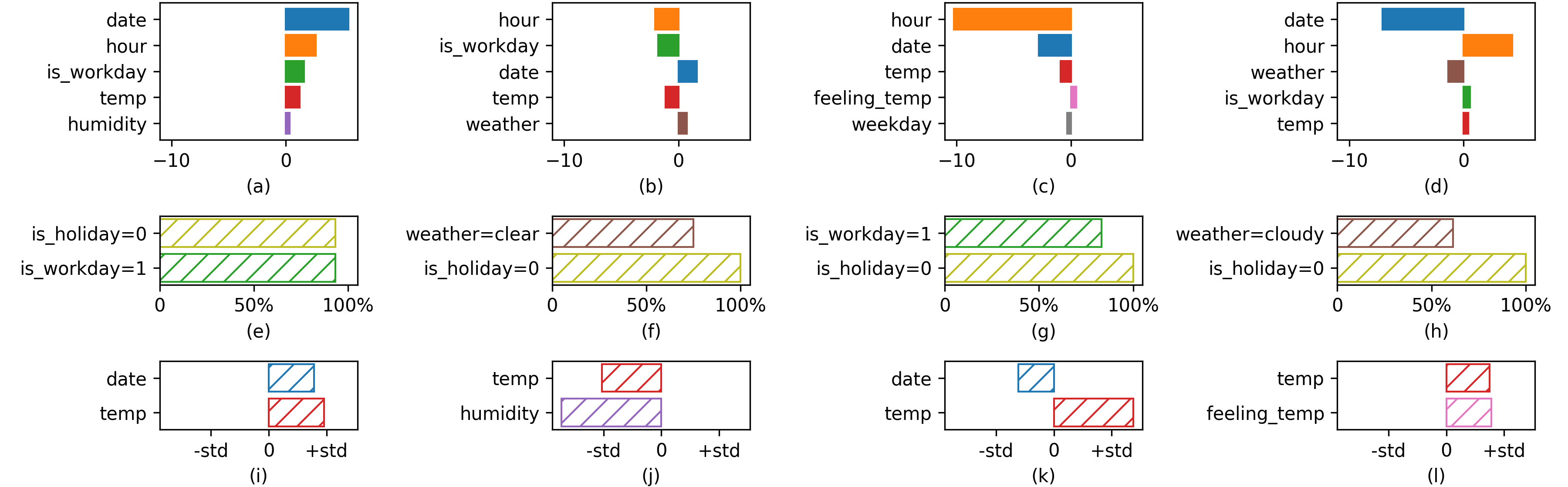}
    \vspace{-2em}
    \caption{Hierarchical explanation.}
  \end{subfigure}
  \caption{Running CohEx on the bike sharing dataset using the baseline methods. Each column represents a cohort. (a-d) The cohort importance of the top 5 features in each cohort; (e-h) the top two most homogeneous categorical features in each cohort, measured in homogeneity; (i-l) the top two most abnormal continuous features. For VINE and REPID, this is measured in the \textit{cohort mean sample}'s different w.r.t. dataset mean in standard deviations; for hierarchical and the CohEx result in the main text, this is measured in the \textit{centroid}'s different w.r.t. dataset mean in standard deviations}
  \label{fig:bike-result-others}
\end{figure*}

Fig.~\ref{fig:bike-result-others} shows the result of running the baseline methods in the bike sharing dataset. Similarly, the first row of each method represents the explanations, and the bottom two rows show the most notable characteristics of each of the cohorts. The second row shows the most abnormal/homogeneous categorical features, and the last row shows the most abnormal continuous features. Note that the meaning of the abnormal continuous features (last row) differs slightly between methods: for CohEx and the hierarchical method, it shows the difference between the \textit{centroid} and the mean of the dataset, while VINE and REPID show the difference between the \textit{cohort average} and the dataset mean, since the latter two methods do not utilize centroids.

The main difference between CohEx and the other methods is that the explanations from the baseline methods are mostly dominated by one feature for each cohort. CohEx is able to amplify the relative importance of features with lower importances, giving a more detailed explanation for each of the cohorts. This further demonstrates CohEx's advantage in its locality.

\subsubsection{Alternative base method}

In Sect.~8, we evaluated CohEx with other baseline methods while using SHAP as the base local explainer. In this section, we provide additional evaluations if we use LIME as the base method. 

Table~\ref{tab:bike-lime} shows the evaluation of the proposed criterion using SHAP. The comparison is mostly in line with the result in the main text.

\begin{table*}[htp]
\small
\centering
\begin{tabular}{@{}cccccc@{}}
\toprule
 & Clustering Penalty ($\downarrow$) & Disjoint Cohort & Cohort Locality ($\downarrow$) & Cohort Stability ($\uparrow$) & Importance Stability ($\downarrow$) \\ \midrule
\multicolumn{6}{c}{\textbf{Bike Sharing (LIME)}} \\ \midrule
\multicolumn{1}{c|}{VINE} & $0.0739 \pm 0.0018$ & $\times$ & $0.1710 \pm 0.0914$ & $0.0047 \pm 0.0299$ & $0.1464 \pm 0.0774$ \\
\multicolumn{1}{c|}{REPID} & $0.0752 \pm 0.0016$ & $\checkmark$ & $0.3619 \pm 0.3256$ & $\mathbf{0.0164 \pm 0.0706}$ & $0.1776 \pm 0.1072$ \\
\multicolumn{1}{c|}{Hierarchical} & $0.0900 \pm 0.0151$ & $\checkmark$ & $1.1659 \pm 0.0170$ & $0.0000 \pm 0.0000$ & $0.1608 \pm 0.0194$ \\
\multicolumn{1}{c|}{CohEx} & $\mathbf{0.0591 \pm 0.0472}$ & $\checkmark$ & $\mathbf{0.1546 \pm 0.1001}$ & $0.0000 \pm 0.0000$ & $\mathbf{0.0622 \pm 0.0199}$  \\ \bottomrule
\end{tabular}
\caption{Evaluating the cohort explanation algorithms in the patient classification task, using vanilla SHAP as the base local explanation method. The ($\downarrow$) symbol after the criteria represents that the lower the better, and ($\uparrow$) represents the higher the better.}
\label{tab:bike-lime}
\end{table*}

\subsection{Digit Classification}
\label{app:mnist}

We also evaluated our algorithm on a vision-based task, that is, digit classification in MNIST \citep{mnist}. Note that the MNIST inputs are cropped, centered, and relatively simple, so measuring the euclidean distance between the input images $x_i$ / the local explanation heatmaps $\omega(X, x_i)$ will be meaningful. This may not be true for more complicated tasks, where semantically similar images/heatmaps may not be close to the raw pixels. For these tasks, an alternative distance measurement should be used instead of the Euclidean distance.

We built the target model as a simple convolutional neural network with two convolutional layers followed by two fully connect layers. The convolutional layers each have a kernel size of 5, and maps to 10 and 20 channels, respectively. Both layers are followed by a max-pooling layer and then a ReLU activation, while the second layer has an additional dropout layer between the convolutional and the pooling layers. The fully connected layers map the final $20 \times 4 \times 4$ pixels to 50 features, then a ReLU activation and a dropout are maps to ten class outputs by another fully connected layer. We trained the model for two epochs with an SGD optimizer and achieved an accuracy of 93.93\% on the test dataset. The model is trained using the full training dataset with all digits.

To demonstrate the properties of cohort explanations, we only include a small set of 200 images of digits 7 and 9 during explanation evaluation. This experiment aims to see if CohEx can focus on pixels that separate these specific digits. Note that the network is trained using the full 10-digits and outputs a probability for each of the classes. We opt to use DeepLIFT \citep{deeplift} as the base explainer, an approximation of SHAP on deep models. DeepLIFT will generate ten feature importance images with respect to all ten classes, even if only two classes are present in the testing dataset. We set the expected number of cohorts to 4, allowing the framework to detect interesting subgroups within each digit class. 

\begin{figure*}[tp]
  \centering
  \begin{subfigure}[b]{0.47\textwidth}
    \centering
    \includegraphics[width=\textwidth]{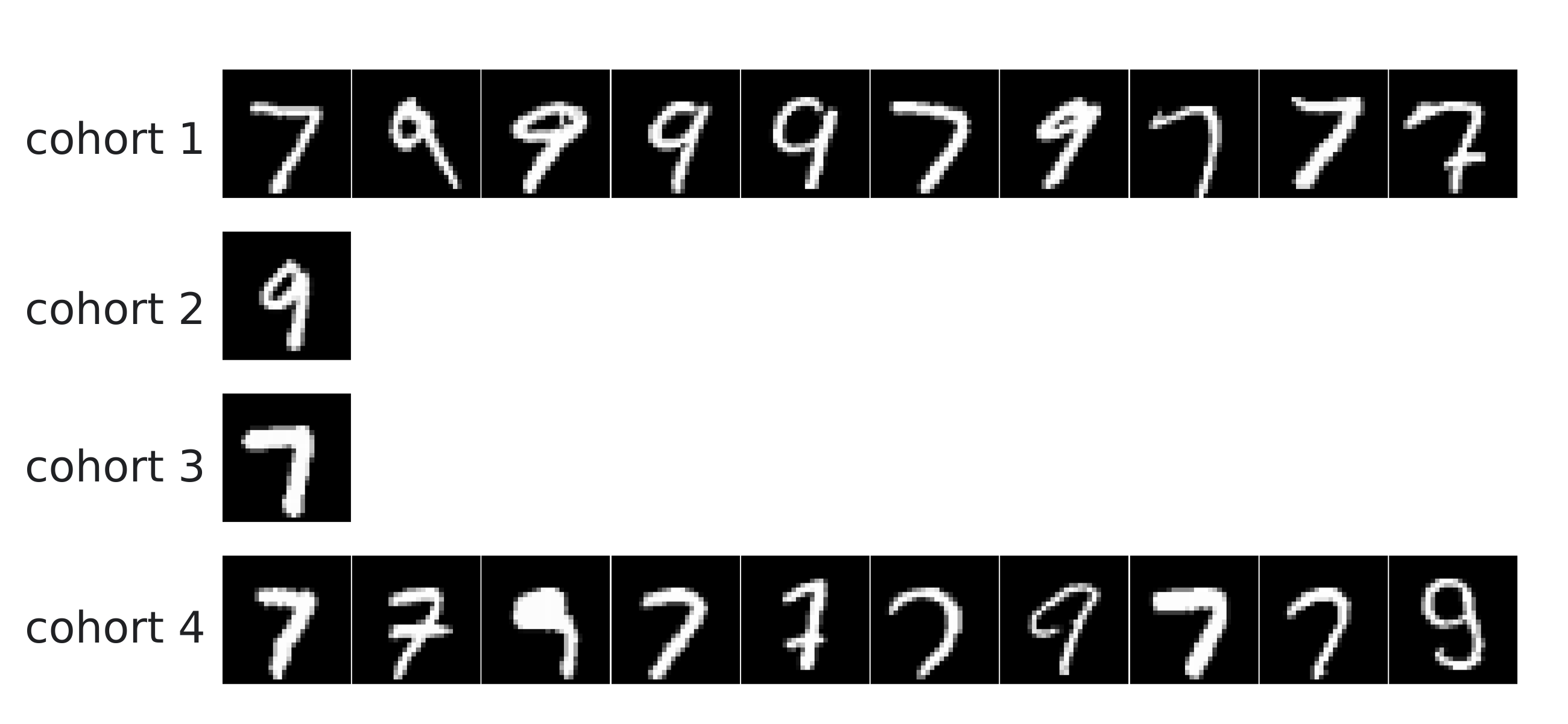}
    \caption{Cohort definition using VINE.}
  \end{subfigure}\quad
  \begin{subfigure}[b]{0.47\textwidth}
    \centering
    \includegraphics[width=0.935\textwidth]{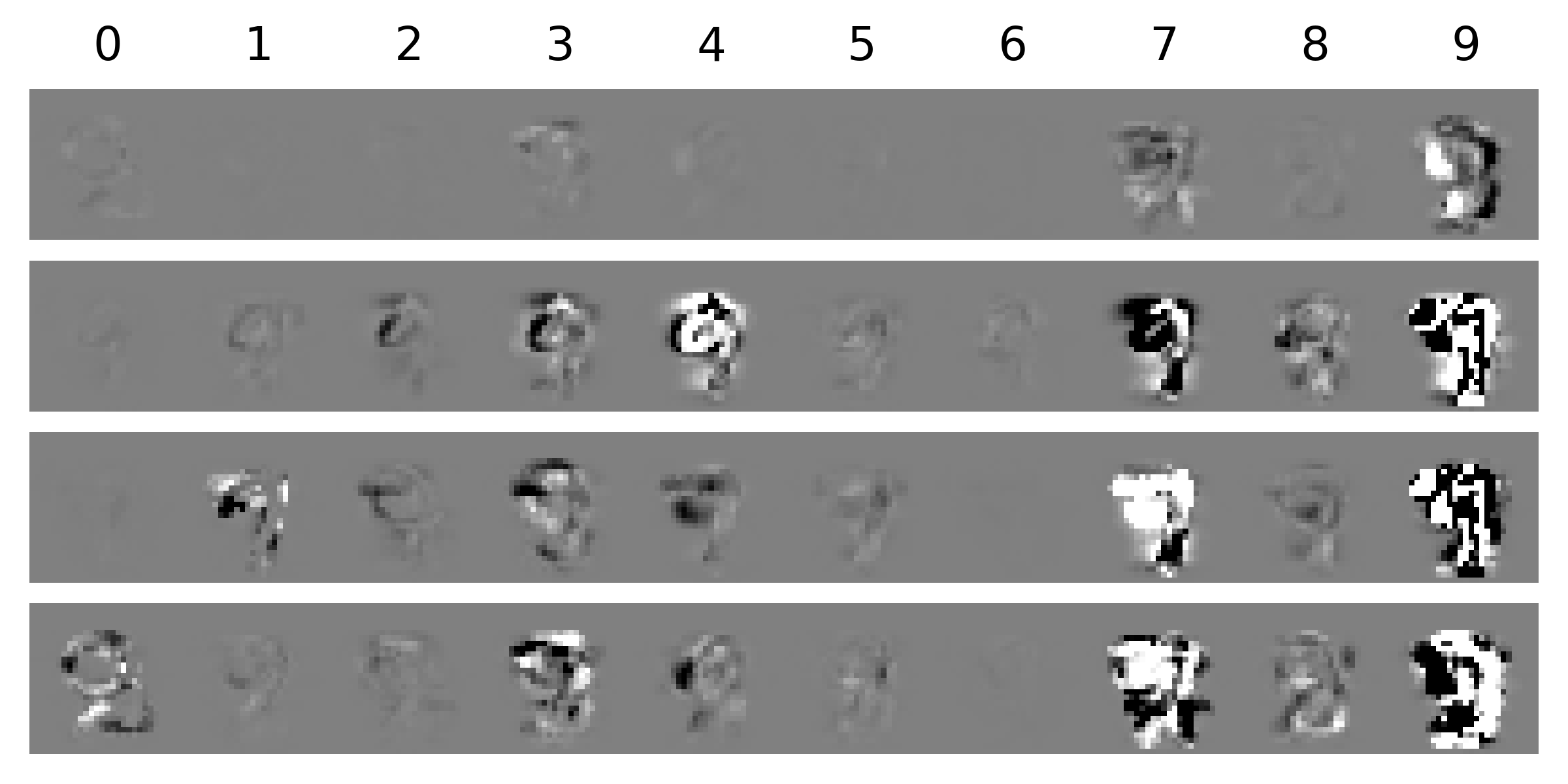}
    \caption{Cohort importance using VINE.}
  \end{subfigure}
  \begin{subfigure}[b]{0.47\textwidth}
    \centering
    \includegraphics[width=\textwidth]{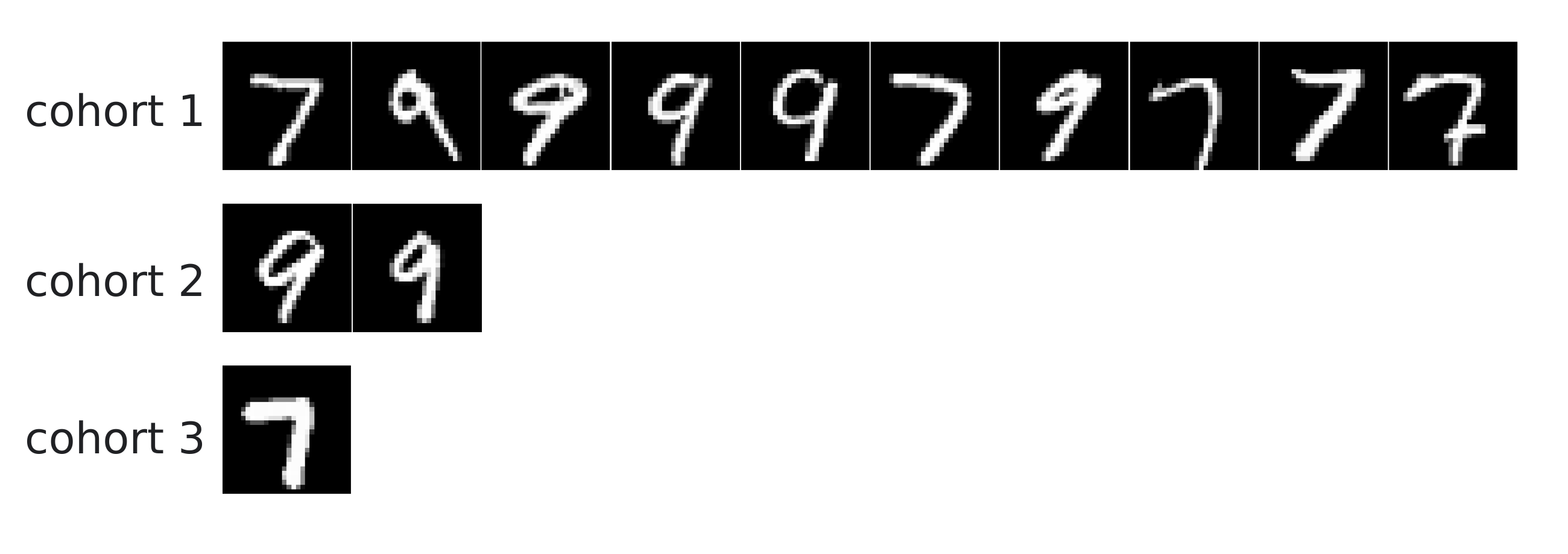}
    \caption{Cohort definition using REPID.}
  \end{subfigure}\quad
  \begin{subfigure}[b]{0.47\textwidth}
    \centering
    \includegraphics[width=0.935\textwidth]{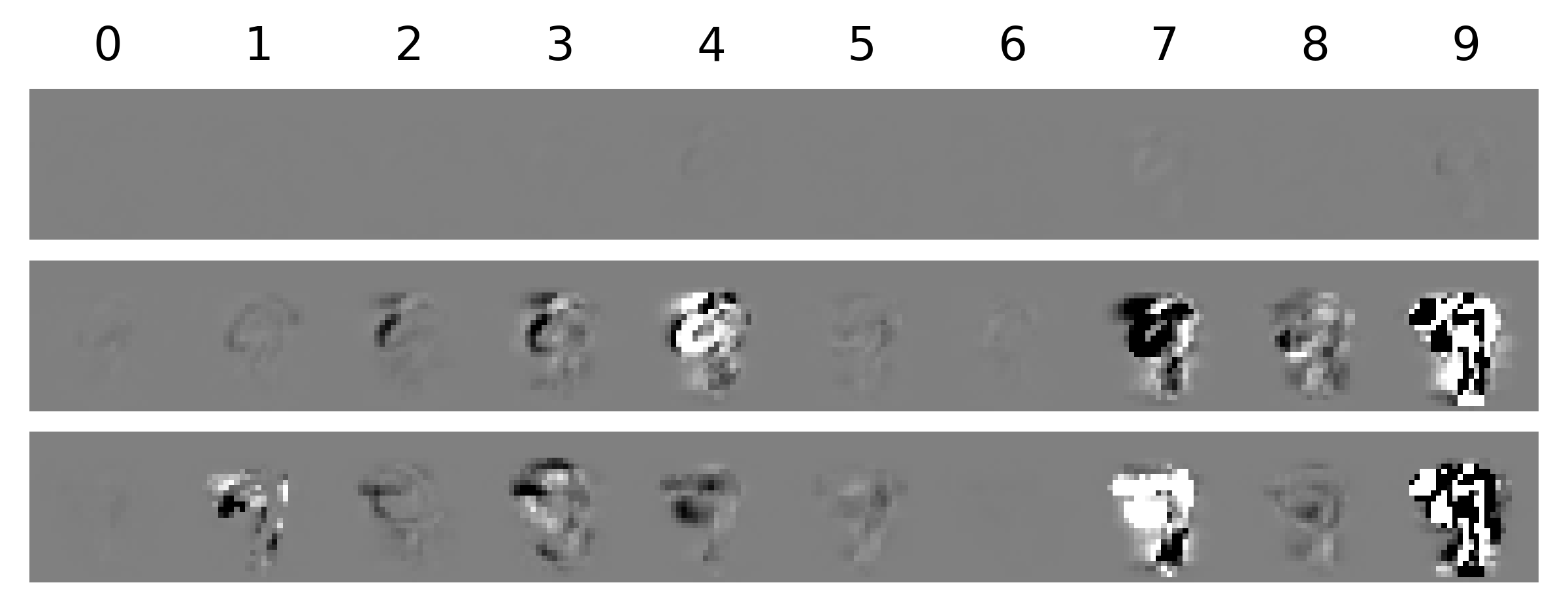}
    \caption{Cohort importance using REPID.}
  \end{subfigure}
  \begin{subfigure}[b]{0.47\textwidth}
    \centering
    \includegraphics[width=\textwidth]{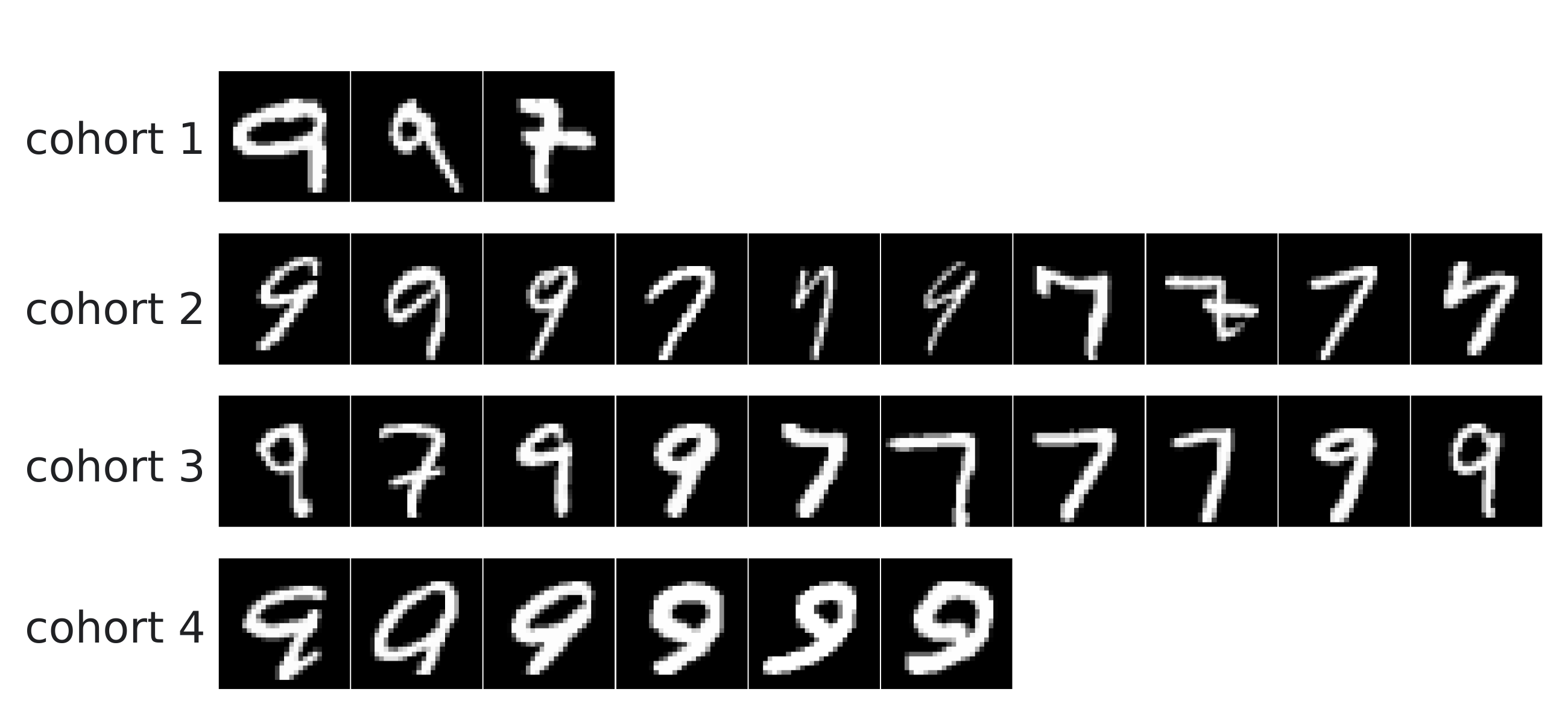}
    \caption{Cohort definition through hierarchical explanation.}
  \end{subfigure}\quad
  \begin{subfigure}[b]{0.47\textwidth}
    \centering
    \includegraphics[width=0.935\textwidth]{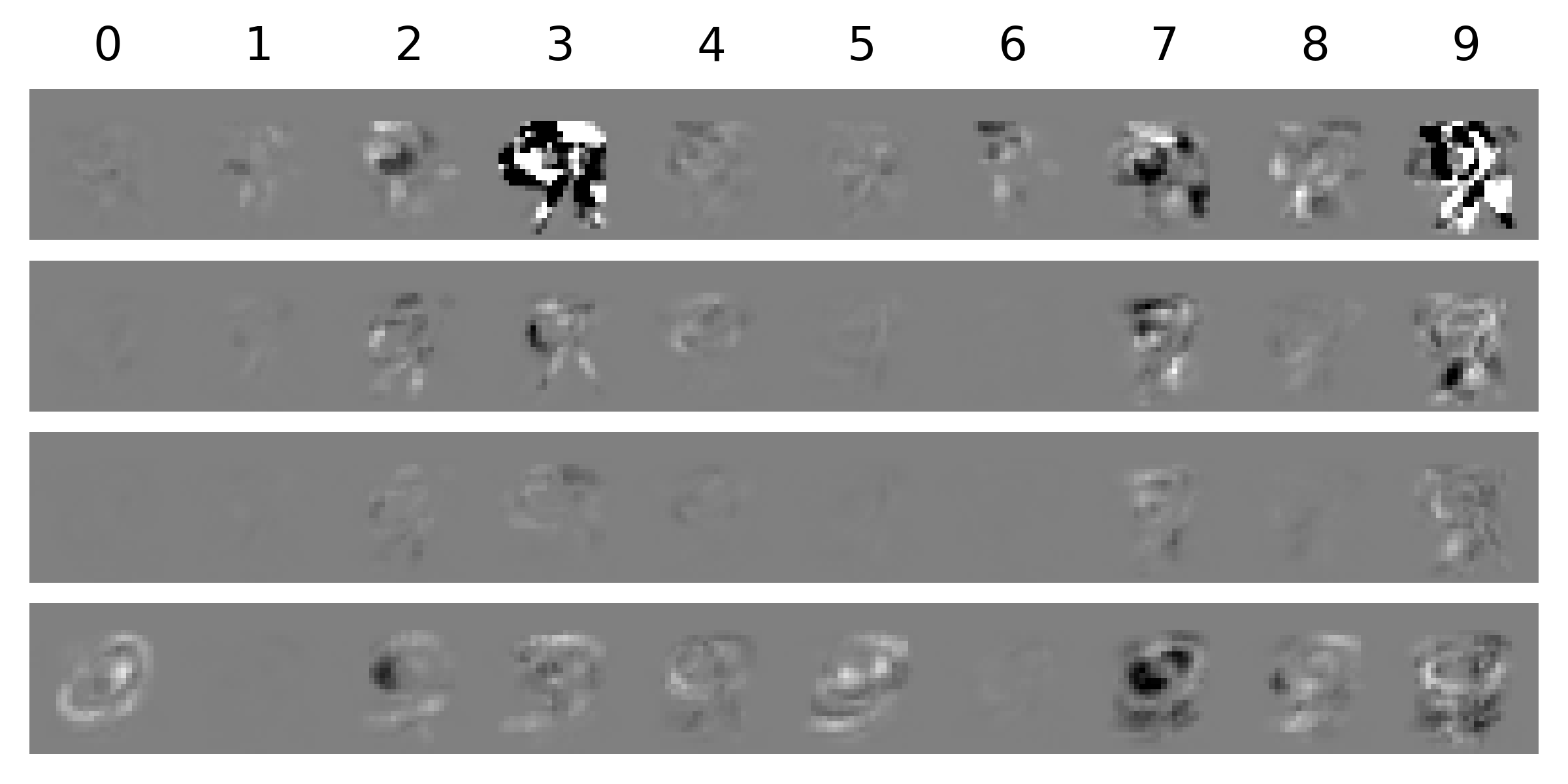}
    \caption{Cohort importance through hierarchical explanation.}
  \end{subfigure}
  \begin{subfigure}[b]{0.47\textwidth}
    \centering
    \includegraphics[width=\textwidth]{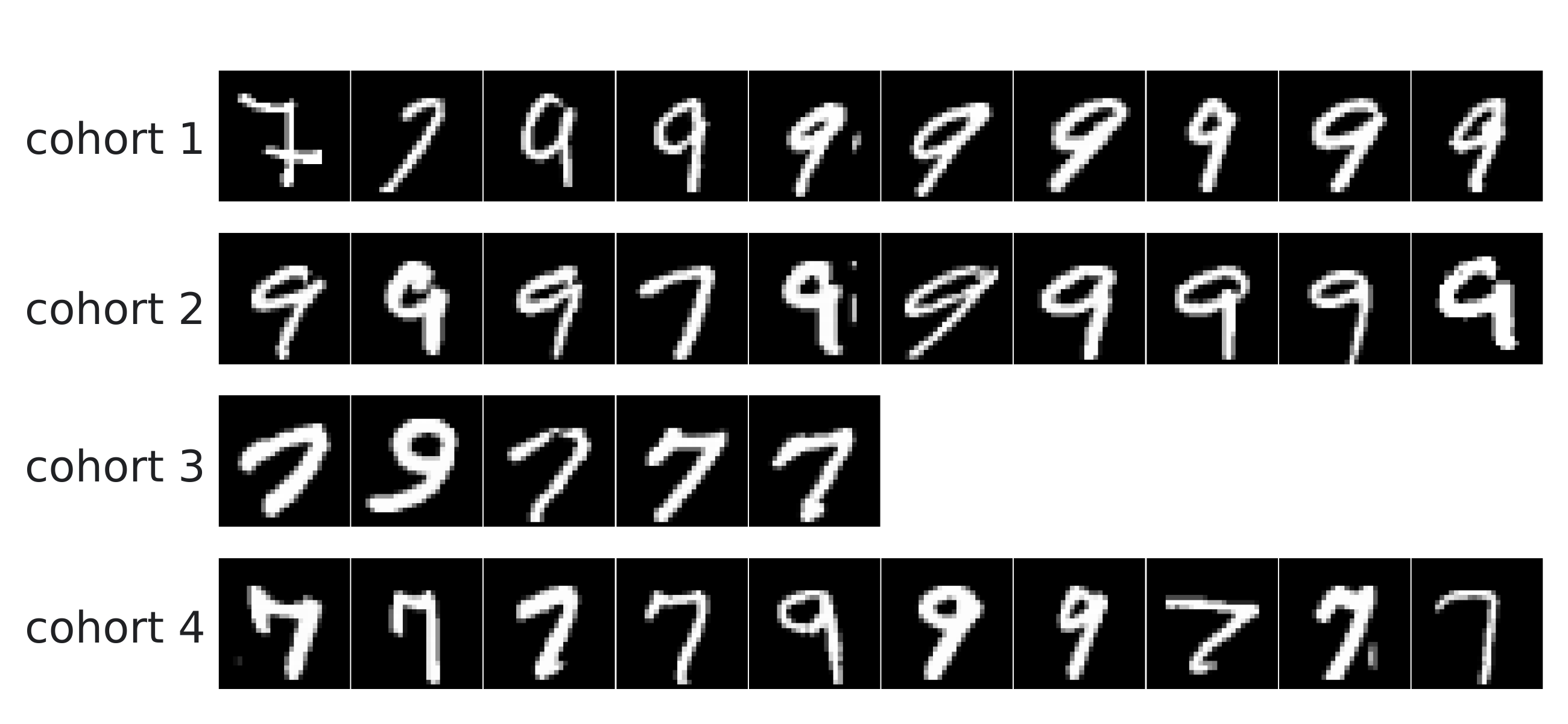}
    \caption{Cohort definition using CohEx.}
  \end{subfigure}\quad
  \begin{subfigure}[b]{0.47\textwidth}
    \centering
    \includegraphics[width=0.935\textwidth]{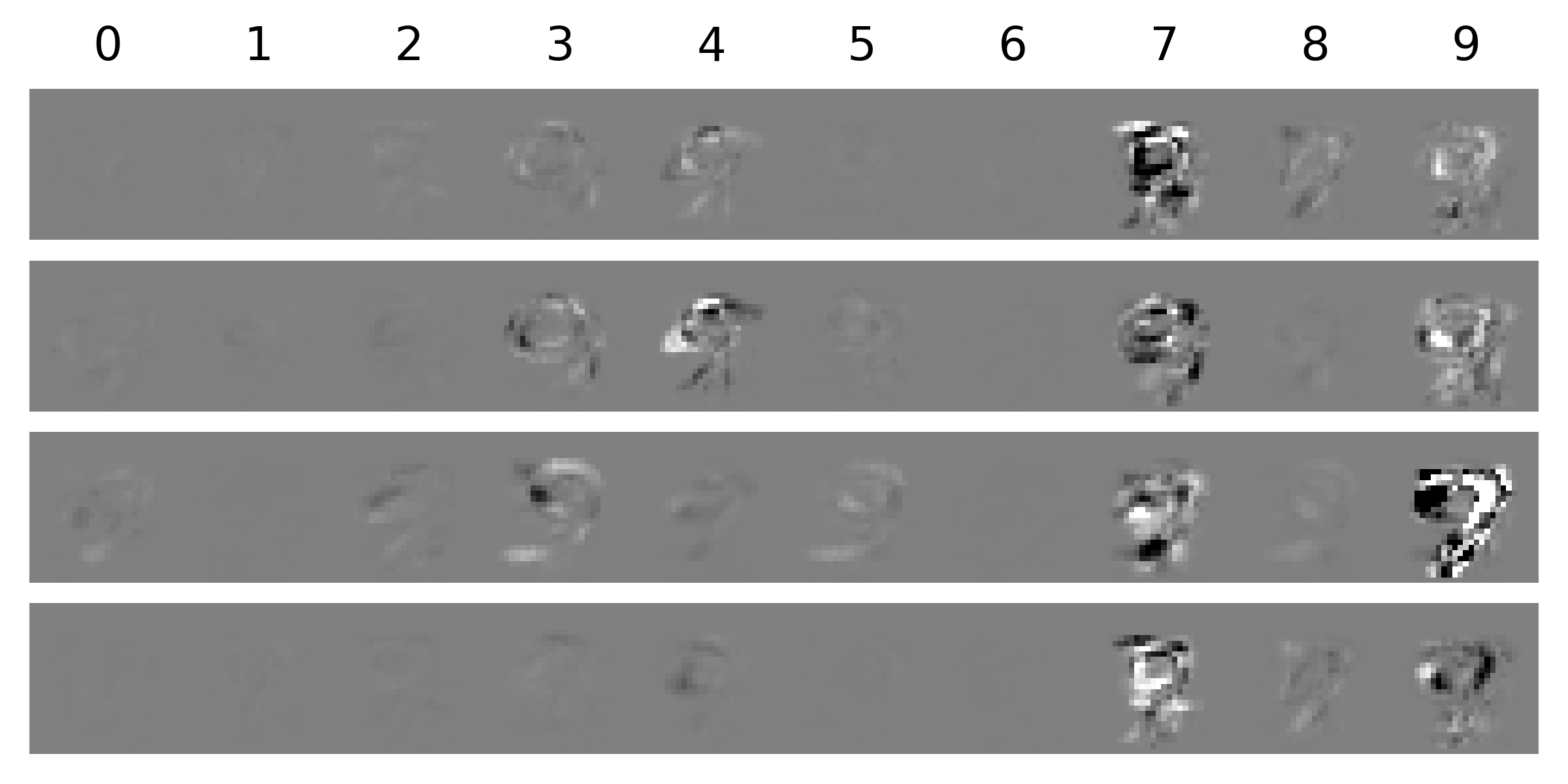}
    \caption{Cohort importance using CohEx.}
  \end{subfigure}

  \caption{Running the four cohort explanation algorithms on the MNIST subset. (a, c, e, g) Definitions of the cohort. Each row represents a cohort, and the images are the first few samples in the cohorts; (b, d, f, h) Each row represents the importance of one cohort. Each column represents the importance of a specific digit, from 0-9. Bright and dark regions denote pixles with high positive/negative importance, respectively, and gray areas have low importance. All importances use the same scale; (a, b) VINE; (c, d) REPID; (e, f) Hierarchical cohort explanation; (g, h) Proposed CohEx framework.}
  \label{fig:mnist-full}
\end{figure*}

Fig.~\ref{fig:mnist-full} shows the result of applying the cohort explanation algorithms in the MNIST subset. The left images show the first 10 members of each cohort, and the left images display the importance w.r.t. each of the ten classes corresponding to the cohorts. Note that clustering on high-dimensional data such as images is a challenging topic, and thus the performance of all methods degrade: all methods yields unbalanced partitions. The VINE's result is the most severe, where it fails to find a meaningful partitioning, and three out of the four cohorts contains very few sample. We note that, though limited by the performance of the clustering algorithm, CohEx's explanation is more specific to this specific task. All other methods contain a nontrivial importance for classes other than 7 and 9, which is what DeepLIFT produce based on the underlying model. However, CohEx's unique importance recomputing step allows it to refine the importances, and reduces the importance for nonexisting classes. This emphasis on locality can be especially crucial for generating explanation that are more specific to downstream tasks on models that are pre-trained on a large dataset or a complicated task. 

\bibliography{cohort-explanation.bib}

\end{document}